\documentclass{article}

\usepackage[final]{neurips_2024}

\usepackage[utf8]{inputenc} 
\usepackage[T1]{fontenc}    
\usepackage{hyperref}       
\usepackage{url}            
\usepackage{booktabs}       
\usepackage{amsfonts}       
\usepackage{nicefrac}       
\usepackage{microtype}      
\usepackage{xcolor}         

\usepackage{amsmath}        
\DeclareMathOperator*{\argmin}{arg\,min}
\usepackage{graphicx}
\newtheorem{lemma}{Lemma}
\newtheorem{theorem}{Theorem}
\usepackage{algorithm}
\usepackage{algorithmic}
\usepackage{tabularx}
\usepackage{comment}
\usepackage{subcaption}

\title{Multi-model Ensemble Conformal Prediction \\
in Dynamic Environments}

\author{%
  Erfan Hajihashemi \\
  Department of Electrical Engineering \& Computer Science\\
  University of California, Irvine \\
  \texttt{ehajihas@uci.edu} \\
   \And
   Yanning Shen\thanks{Corresponding author} \\
   Department of Electrical Engineering \& Computer Science\\
   University of California, Irvine \\
   \texttt{yannings@uci.edu} \\
}

\begin{document}
\maketitle
\begin{abstract}
Conformal prediction is an uncertainty quantification method that constructs a prediction set for a previously unseen datum, ensuring the true label is included with a predetermined coverage probability. Adaptive conformal prediction has been developed to address data distribution shifts in dynamic environments. However, the efficiency of prediction sets varies depending on the learning model used. Employing a single fixed model may not consistently offer the best performance in dynamic environments with unknown data distribution shifts. To address this issue, we introduce a novel adaptive conformal prediction framework, where the model used for creating prediction sets is selected ‘on the fly’ from multiple candidate models. The proposed algorithm is proven to achieve strongly adaptive regret over all intervals while maintaining valid coverage. Experiments on real and synthetic datasets corroborate that the proposed approach consistently yields more efficient prediction sets while maintaining valid coverage, outperforming alternative methods.
\end{abstract}
\section{Introduction}
Most machine learning algorithm designs aim to enhance label prediction accuracy. Nevertheless, a significant challenge persists as many models demonstrate limitations in predicting labels with high certainty, falling short of achieving the desired levels of accuracy and other critical evaluation metrics. In applications such as medical diagnosis, it is sometimes more efficient to predict a subset of labels rather than a single label \citep{levy2021,straitouri2023}. This necessitates predicting a set of candidate labels with a valid coverage probability, rather than limiting to a single label. \par
One of the most widely used frameworks for set prediction is conformal prediction \citep{vovk2005}. Conventional conformal prediction algorithms can achieve the desired coverage assuming the exchangeability of data \citep{balasubramanian2014}. However, in many real-world online problems, the distribution of data shifts over time, making the exchangeability assumption no longer applicable. Consequently, adaptive conformal prediction algorithms have been developed \citep{gibbs2021}, where prediction sets are constructed in a time-varying manner. Despite these advancements, the efficiency (e.g., prediction set size or regret) of previous conformal prediction methods in online settings with distribution shifts heavily depends on the model employed, and a single model may not consistently perform well across various distribution shifts. Creating an efficient prediction set size is as important as obtaining the desired coverage. Trivial cases may arise where the prediction set alternates between an empty set and the full set of labels, with the full set being created with a probability equal to the desired coverage probability, and empty sets otherwise. This approach achieves the desired coverage but leads to impractical prediction sets \citep{bhatnagar2023}. To address this limitation, our proposed algorithm incorporates multiple learning models simultaneously. It dynamically selects the suitable model based on the performance of each model with the most recently received data. \par
\noindent \textbf{Related work:}
Conformal prediction \citep{vovk2005, shafer2008, vovk2015} is an effective method for uncertainty quantification that has been widely used to predict a set of candidate labels for income data. It treats the learning model as a black box and provides a prediction set for new test data. Conformal prediction is applicable in both Classification \citep{ding2024, shi2013, romano2020} and Regression \citep{romano2019, papadopoulos2011, bostrom2017} problems. In dynamic environments where data distribution shifts over time, using vanilla conformal prediction algorithms may not lead to the desired coverage performance. To cope with this challenge, conformal prediction in dynamic environments has been studied recently. \citep{tibshirani2019} explored conformal prediction for dynamic settings using a reweighting approach, but their method requires prior information about the data dependency structure; \citep{barber2023} resolved this dependency by requiring weights to be fixed. Incorporating time-varying coverage probability was introduced in \citep{gibbs2021}, but determining the appropriate step size remains a significant challenge.  One way to address dynamic environments is adopting learning with expert advice \citep{cesa1997, vovk1995, littlestone1994}. \citep{zaffran2022} suggested that tuning the step size based on expert (base learner) aggregation could be an effective strategy; however, this method causes each expert to receive equal impact from all historical data, which makes the algorithm unable to adapt to sharp distribution shifts. Furthermore, \citep{gibbs2022} introduced an approach that employs multiple experts, each assigned a distinct step size from a pool of candidate sizes, resulting in varying coverage probabilities at each time $t$. While \citep{gibbs2022} successfully demonstrated adaptive regret across time intervals of a fixed width, \citep{bhatnagar2023} points out that this approach fails to achieve suitable regret across varying widths of arbitrary time intervals simultaneously. \citep{bhatnagar2023} addressed this limitation by establishing strongly adaptive regret \citep{daniely2015} across any arbitrary time interval width. Their proposed methodology assigns a specific time interval to each expert. Despite achieving sublinear strongly adaptive regret, the efficacy of this approach depends on the hyper-parameter selection that determines each expert's lifetime. The proposed method in our study also employs experts who operate within specific time intervals. However, each expert consists of multiple learning models, aiming to select the appropriate model according to the specific data distribution during its operation, resulting in more efficient prediction sets.\par
\textbf{Contributions.} Overall, our contributions can be summarized as follows: \\
\textbf{\MakeUppercase{\romannumeral 1})} We introduce a novel adaptive conformal prediction algorithm, \textbf{S}trongly \textbf{A}daptive \textbf{M}ultimodel Ensemble \textbf{O}nline \textbf{C}onformal \textbf{P}rediction (SAMOCP), designed for dynamic environments with unknown distribution shifts. This algorithm incorporates multiple models and dynamically selects a model based on its performance in previous time steps. \\
\textbf{\MakeUppercase{\romannumeral 2})} We demonstrate that SAMOCP exhibits strongly adaptive regret for any arbitrary time interval while ensuring valid coverage.\\
\textbf{\MakeUppercase{\romannumeral 3})} Through experimental tests on classification tasks subject to distribution shifts, we demonstrate that SAMOCP outperforms existing methods by constructing more efficient prediction sets while also achieving a coverage probability closely aligned with the target value.
\section{Preliminaries}
\label{sec:Preliminaries and Problem Statement}
 This section explains standard conformal prediction and adaptive online conformal predictions, where data is collected sequentially. We begin with outlining standard conformal prediction. Given a miss coverage probability \(\alpha\), a learning model \(m\) , a historical dataset \(\{(X_{\tau}, Y_{\tau}^{\text{true}})\}_{\tau=1}^{t-1}\), and a new data  \(X_t \in \mathcal{X}\), the objective is to construct a prediction set \(C_{\alpha}^{m} (X_t) \subseteq \mathcal{Y}:= \{1, 2, \ldots, K\} \), where \(K\) denotes the total number of classes, such that \(C_{\alpha}^{m} (X_t)\) contains the true label \(Y_t^{\text{true}}\) with probability \(1 - \alpha\). In the online setting, historical dataset is updated to \(\{(X_{\tau}, Y_{\tau}^{\text{true}})\}_{\tau=1}^{t}\) at the end of each time $t$ when true label \(Y_t^{true}\) for input data $X_t$ is observed. In this scenario, conformal prediction treats the historical dataset as a calibration dataset, which is utilized to determine whether a candidate label \(Y \in \mathcal{Y}\) should be included in the prediction set. Consequently, in an online manner, conformal prediction relies on an evolving calibration dataset, which contains all the historical data ‘on the fly’ to decide which candidate labels should be included in the prediction set.  Non-conformity scores  \(\{S^{m}(X_{\tau}, Y_{\tau}^{\text{true}})\}_{\tau=1}^{t-1}\) are introduced. Specifically, each non-conformity score \(S^{m}(X_{\tau}, Y_{\tau}^{\text{true}})\)  assesses the disagreement between the ground-truth label \(Y_{\tau}^{\text{true}}\) and predicted label \(\hat{f}^{m}(X_{\tau})\). Upon obtaining a new datum \(X_{t}\), the standard conformal prediction algorithm constructs the prediction set for $X_t$ as \( C_{\alpha}^{m} (X_t) = \{Y \in \mathcal{Y} \mid S^{m}(X_t, Y) \le \hat{q}_{\alpha}^{m} \}\),  where the threshold \(\hat{q}_{\alpha}^{m}\) is obtained as
\begin{equation}
  \hat{q}_{\alpha}^{m} = Quantile \left(\frac{\lceil t (1-\alpha) \rceil}{t-1},\{S^{m}(X_{\tau},Y_{\tau}^{true})\}_{\tau=1}^{t-1}\right).
  \label{eq:quantile_formula}
\end{equation}
 The Quantile function sorts all non-conformity scores of the historical data and then identifies the \(\lceil t (1-\alpha) \rceil\)th smallest score as \(\hat{q}_{\alpha}^{m}\). Note that  \(1 - \alpha\) is fixed, hence conformal prediction cannot readily cope with potential data distribution shifts in dynamic environments. Adaptive conformal prediction algorithms have been developed to address this issue, allowing the miss coverage probability to vary at each time $t$ and thereby enabling the algorithm to dynamically adapt to potential shifts in the distribution. In such a scenario, \(\hat{q}_{\alpha}^{m}\) can be obtained by replacing \(\alpha\) with \(\alpha_t\) in \eqref{eq:quantile_formula}. Then at each time step \(t\), $\alpha_t$ is updated after observing \(Y_t^{true}\).\par Recent studies on online conformal prediction with distribution shifts have incorporated adaptive miss coverage probabilities to address dynamic environments. However, methods based on a single learning model may not achieve consistently reliable performance in dynamic environments. This underscores the necessity for employing multiple models and adaptive strategies to determine the appropriate model for each time $t$. To this end, in this work, we introduce a novel adaptive multi-model online conformal prediction algorithm designed to identify the suitable learning model at each time \(t\) within dynamic environments. At time slot \(t\), the goal is to construct a prediction set for the new data \(X_t\), based on the historical dataset \(\{(X_{\tau}, Y_{\tau}^{\text{true}})\}_{\tau=1}^{t-1}\), such that the true label is included in the prediction set with probability \(1-\alpha\). The proposed algorithm for dynamic settings achieves strongly adaptive regret while ensuring valid coverage.
\section{Methodology}
In this section, two adaptive algorithms are developed for static and dynamic environments respectively. Subsection \ref{sec:model-selection} develops the \textbf{M}ultimodel Ensemble  \textbf{O}nline \textbf{C}onformal \textbf{P}rediction (MOCP) algorithm to identify the suitable learning model among $M$ distinct candidates in a static environment. Subsequently, in Subsection \ref{sec:expert-selection}, we propose SAMOCP, an adaptation of MOCP tailored for dynamic environments with unknown distribution shifts.
\subsection{Multi-model Conformal Prediction in Static Environments}
\label{sec:model-selection}
Note that the non-conformity score \(S^{m}(X_{\tau},Y_{\tau}^{true})\) depends on the learning model. Such dependency leads to a model-specific ordering of non-conformity scores, yielding different prediction sets for each model. After observing \(Y_t^{true}\), the adaptive miss coverage probability \(\alpha_t\) must be updated for time \(t+1\) to cope with distribution shifts effectively. Given that different models achieve different prediction sets, assigning and updating the same \(\alpha_t\) for different models would be inadequate. Instead, at each time $t$, we assign a specific miss coverage probability to each model \(m \in [M]\), denoted as \(\alpha_t^{m}\), and update it based on the corresponding prediction set. Consequently, for $M$ learning models, there are \(M\) candidates for miss coverage probability \(\alpha_t\) at each time $t$. Each candidate is updated according to a distinct rule. These \(M\) update rules operate in parallel, with each one updating the corresponding miss coverage probability upon observing the true label. Next, the update procedure for \(\alpha_t^{m}\) will be examined, followed by a detailed explanation of how each instance of MOCP selects the appropriate miss coverage probability from \(M\) distinct options at each time step. \\ 
To update miss coverage probability \(\alpha_t^{m}\), we adopt the pinball loss \citep{koenker1978}, which can be written as
\begin{equation}
  L(\Bar{\alpha}_t^m, \alpha_t^m) = \alpha (\Bar{\alpha}_t^m - \alpha_t^m) - \min \{0 , \Bar{\alpha}_t^m - \alpha_t^m\},
  \label{eq:loss-model-m}
\end{equation}
where 
\begin{equation}
  \Bar{\alpha}_t^m  = \sup\{\Tilde{\alpha} : Y_t^{true} \in C_{\Tilde{\alpha}}^m (X_t)\}
  \label{eq:optimum-alpha}
\end{equation}
  is the best possible value of miss coverage probability for model $m$ at time $t$ which constructs the smallest prediction set that covers $Y_t^{true}$.The miss coverage probability $\alpha_{t+1}^m$ can be updated via SF-OGD  \citep{orabona2018} as
\begin{equation}
  \alpha_{t+1}^{m} = \alpha_t^m - \eta \frac{\nabla_{\alpha_t^m} L(\Bar{\alpha}_t^m, \alpha_t^m)}{\sqrt{ \sum_{\tau=1}^{t} \|\nabla_{\alpha_{\tau}^m} L(\Bar{\alpha}_{\tau}^m, \alpha_{\tau}^m)\|_2^2 }},
  \label{eq:update-rule}
\end{equation}
where $\eta$ is the learning rate and
\begin{equation}
  \nabla_{\alpha_t^m} L(\Bar{\alpha}_t^m, \alpha_t^m) = \mathbb{I} [\Bar{\alpha}_t^m < \alpha_t^m ] - \alpha = err_t^{m} - \alpha,
  \label{eq:grad}
\end{equation}
with  \(err_t^{m}:= \mathbb{I}[Y_t^{true} \notin C_{\alpha_t^m}^m]\) equals $1$ if the predicted set does not contain the true label \(Y_t^{true}\), and $0$ otherwise. According to the updating rule outlined in equation \eqref{eq:update-rule}, the adjustment of $\alpha_t^m$ at each time $t$  is governed by the \(\nabla_{\alpha_t^m} L(\Bar{\alpha}_t^m, \alpha_t^m)\), as detailed in equation  \eqref{eq:grad}. When \(err_t^{m} =1\), it signals that the coverage probability \(1 - \alpha_t^m\) is too small, resulting in a prediction set that can not encompass \(Y_t^{true}\). Consequently, there's a necessity to enlarge the coverage probability, effectively achieved by reducing \(\alpha_t^m\), which would be facilitated by \eqref{eq:update-rule}; given that the denominator in the second term is always positive and the gradient will be positive in this scenario. On the other hand, when \(\Bar{\alpha}_t^m > \alpha_t^m\), \(1 - \alpha_t^m\) leads to a prediction set that covers \(Y_t^{true}\)  but also includes unnecessary labels \(\mathcal{Y}':= \{Y' \in \mathcal{Y} \mid \hat{q}_{\Bar{\alpha}_t^m}^m < S^{m}(X_t, Y') \le \hat{q}_{\alpha_t^m}^m\}\). In such cases, optimization necessitates increasing  \(\alpha_t^m\) to avoid including unnecessary labels and output a more efficient prediction set. This adjustment is facilitated by the update rule \eqref{eq:update-rule}. \par
Additionally, the weight \(w_t^{m}\) is assigned to each model \(m \in [M]\), which influences the selection of its corresponding miss coverage probability \(\alpha_t^m\). MOCP learns which model to select over time based on the performance of each model over previous time steps, as reflected in \(w_t^{m}\). The algorithm updates the weight associated with each model after revealing the true label \(Y_t^{true}\). This update is performed with respect to the loss function of the corresponding miss coverage probability. Specifically, \(w_t^{m}\) can be updated by
\begin{equation}
  w_{t+1}^{m} = w_t^{m} \exp\left(-\epsilon L\left(\Bar{\alpha}_t^m, \alpha_t^m\right)\right),
  \label{eq:update-rule-model-weight}
\end{equation}
where \(0< \epsilon < 1\) is the step size. At each time $t$, upon receiving new data \(X_t\), MOCP first calculates the normalized weights, denoted as  \(\{\Bar{w}_t^{m}\}_{m=1}^M\). For any \(m \in [M]\), \(\bar{w}_t^{m} = \frac{w_t^{m}}{\sum_{j=1}^{M} w_t^{j}}\) ensures that \(\bar{w}_t^{m}  \in [0, 1]\) and represents the likelihood of selecting miss coverage probability  \(\alpha_t^m\). Then, the algorithm selects miss coverage probability \(\alpha^{\hat{m}}_t\), where \(\hat{m} \in [M]\), according to PMF \(\boldsymbol{\Bar{w}}_t:= (\Bar{w}_t^{m})_{m=1}^M\), i.e., each miss coverage probability \(\alpha_t^m\) is selected with probability proportional to the corresponding normalized weight \(\Bar{w}_t^{m}\). The prediction set for \(X_t\) is constructed according to the threshold in \eqref{eq:quantile_formula}, by replacing  \(\alpha\) and \(m\) with \(\alpha^{\hat{m}}_t\) and \(\hat{m}\) respectively. 
After receiving $Y_t^{true}$, each weight \(w_t^{m}\) and miss coverage probability \(\alpha_t^{m}\) are updated according to \eqref{eq:update-rule-model-weight} and \eqref{eq:update-rule}, respectively. This entire process is detailed in Algorithm \ref{alg:alg1-model-selection}. {The MOCP algorithm achieves a runtime of \(\mathcal{O}(T)\) when the number of models $M$ is constant.}

Given that the environment is static, there exists a miss coverage probability that can minimize the loss function for each model $m \in [M]$ over $[T]$, denoted as $\alpha^m$. The best miss coverage probability among $\{\alpha^m\}_{m=1}^M$ can be obtained by
\begin{equation}
  \alpha^{m^*} = \argmin_{\{\alpha^m, m \in [M]\}} \sum_{t=1}^{T} L(\Bar{\alpha}_t^{m}, \alpha^{m})   \hspace{2mm}  \text{with} \hspace{2mm} \alpha^{m} = \argmin_{\alpha_t^{m}} \sum_{t=1}^{T} L(\Bar{\alpha}_t^{m}, \alpha_t^{m}).
  \label{eq:regret-mocp-optimals}
\end{equation}
The following theorem demonstrates that MOCP achieves sublinear regret (See proof in \ref{pr:regret-MOCP-proof}).

\begin{algorithm}
\caption{Multi-model Ensemble Online Conformal Prediction (MOCP)}\label{alg:alg1-model-selection}
\begin{algorithmic}
\REQUIRE $\alpha \in [0,1]$, learning rate $\eta \ge 0$, and step size $\epsilon \in (0, 1)$.
\STATE $\text{$w_1^{m}$} \gets \text{$\frac{1}{M}$}$ \COMMENT{Initialization of weight for each model}
\STATE $\text{$\alpha_1^m$} \gets \alpha$ \COMMENT{Initialization of miss coverage probability for each model}
\FOR{$t \in [T]$}
    \STATE Observe $X_t \in \mathcal{X}$.
    \STATE Calculate normalized weights by $\Bar{w}_t^{m} = \frac{w_t^{m}}{\sum_{j=1}^{M} w_t^{j}}$, $\forall m \in [M]$.
    \STATE Select one of the miss coverage probabilities  $\{\alpha_t^{m}\}_{m=1}^M$ according to PMF  $\boldsymbol{\Bar{w}}_t = (\Bar{w}_t^{m})_{m=1}^M$.
    \STATE Observe true label $Y_t^{true}$ and compute optimal value $\Bar{\alpha}_t^m$ $\forall m \in [M]$ with \eqref{eq:optimum-alpha}.
    \FOR{$m \in [M]$}
        \STATE Obtain loss $L(\Bar{\alpha}_t^m, \alpha_t^{m})$.
        \STATE Update $w_{t+1}^{m}$ with \eqref{eq:update-rule-model-weight}.
        \STATE Update $\alpha_{t+1}^{m}$ with \eqref{eq:update-rule}.
    \ENDFOR
\ENDFOR
\end{algorithmic}
\end{algorithm}

\begin{theorem}\label{thm:regret-alg1}

Algorithm \ref{alg:alg1-model-selection} achieves the following regret bound in a static environment
\begin{equation}
\sum_{t=1}^{T} \sum_{m=1}^{M} \Bar{w}_t^{m}L(\Bar{\alpha}_t^m, \alpha_t^{m}) -  \sum_{t=1}^{T} L(\Bar{\alpha}_t^{m^*}, \alpha^{m^*}) \le 
\sqrt{T}\left(\frac{(1+2\eta)^2}{2\eta} + \frac{\eta}{2\alpha} + \ln{M} + (1+\eta)^2\right). 
  \label{eq:regret-mocp}
\end{equation}
\end{theorem}

\subsection{Multi-model Ensemble Conformal Prediction In Dynamic Environments}
\label{sec:expert-selection}

Through Algorithm \ref{alg:alg1-model-selection}, we demonstrated how to select the suitable miss coverage probability of a model that has achieved lower loss compared to other models over previous time steps. However, this approach assumes a static environment that does not change over time, which may limit the algorithm's effectiveness in dynamic environments where data distribution shifts occur. In addition, the selection of stepsize \(\epsilon\)  in \eqref{eq:update-rule-model-weight} critically affects the performance. In environments with unknown distribution shifts, a large \(\epsilon\) indicates faster adaptation to abrupt changes, whereas a small \(\epsilon\) is more suitable for environments with less variability. Thus, the efficient choice of \(\epsilon\) depends on the variability of the environment, which poses a challenge in scenarios with unknown distribution shifts. \par
\begin{figure}
  \centering
  \includegraphics[width=12cm, height = 4cm]{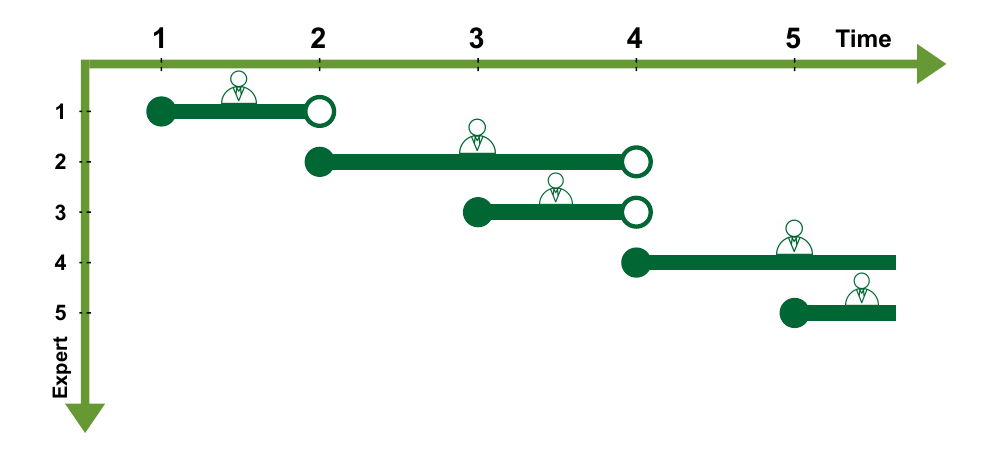}
  \caption{Expert creation over 5 time steps using lifetime formula \eqref{eq:lifetime} when $g = 1$. At each time $t$, an expert is created, marked by a filled circle to indicate the start of the activity, and an unfilled circle to denote the end of the expert's activity. 
  }
  \label{fg:lifetime}
\end{figure}

 
To address this limitation, we introduce Strongly Adaptive (SA)MOCP. Specifically, each instance of MOCP is treated as an \textquoteleft expert\textquoteright, and multiple experts are created at distinct time steps with specific step sizes and lifetimes to cope with potential distribution shifts. At the end of its lifetime, the expert becomes inactive, ensuring that it no longer affects the decision-making process, refer to Figure \ref{fg:lifetime} for an illustration. This strategy prevents outdated experts from contributing to the selection of the suitable miss coverage probability in dynamic environments. Subsequently, SAMOCP dynamically selects the appropriate expert for each time \(t\) and utilizes its chosen miss coverage probability to construct the prediction set. The lifetime of each expert is determined by the specific time \(t\) at which it was created and hyperparameter \(g \in \mathbb{N}\), as \citep{bhatnagar2023}.
\begin{equation}
  \lambda(t) = g \cdot \max_{n \in \mathbb{Z}} \left\{2^n : t \equiv 0 \ \bmod \ 2^n \right\}.
  \label{eq:lifetime}
\end{equation}
The active interval of the expert created at time \(t\) is defined as \([t, t + \lambda(t) - 1]\). Consequently, experts considered active at any given time \(\tau\) are those whose active intervals include \(\tau\). In a dynamic setting with unknown distribution shifts, the best model may vary across different distributions. This variability results in scenarios where, at each time $t$, some active experts might not have adapted to the current data distribution yet, and thus they rely on different models compared to more recently established active experts. In such cases, the miss coverage probability of model \(\hat{m} \in [M]\) chosen by expert \(n\) at time \(t\) is represented as \(\alpha_t^{\hat{m}n}\), and its best possible value is denoted as \(\Bar{\alpha}_t^{\hat{m}n}\). To select the suitable miss coverage probability among different experts, we assigned distinct weights to each, with specific initialization and step sizes for updates. Specifically, this weight for expert \(n\) at time \(t\) is denoted as $h_t^n$, and its step size is defined as \(\epsilon^n := \min (\epsilon, \frac{\sigma}{\sqrt{\lambda(n)}})\), where \(\sigma > 1\) is a constant and \(\lambda(n)\) is the lifetime for expert $n$ that obtained by \eqref{eq:lifetime}. Given that \(n\)th expert is activated at $t = n$, the initialization and update rule for \(h_t^n\) is as follows:
\begin{equation}
   h_{t+1}^{n} = 
   \begin{cases} 
   \epsilon^n & \text{if } t = n - 1 \\
   h_t^n \exp\left(-\epsilon^n \cdot r_t^n\right) & \text{if } t \in [n, n+\lambda(n)-1) \\
   0, & \text{otherwise}
   \end{cases}
  \label{eq:updateh}
\end{equation}
where $r_t^n = L(\Bar{\alpha}_t^{\hat{m}{\hat{n}}}, \alpha_t^{\hat{m}{\hat{n}}}) -  L(\Bar{\alpha}_t^{\hat{m}n}, \alpha_t^{\hat{m}n})$ represents the loss of the $n$th expert relative to the loss of the learner who selects expert $\hat{n}$. We denote the set of active experts at each time $t$ as ${\cal A}(t)$. The learner selects the suitable miss coverage probability among all active experts according to the PMF $\boldsymbol{\Bar{h}}_t:= (\Bar{h}_{t}^n)_{n \in {\cal A}(t)}$, where each $\Bar{h}_t^n$ represents the normalized version of the weight $h_{t}^{n}$, calculated as $\Bar{h}_t^{n} = \frac{h_t^{n}}{\sum_{i \in {\cal A}(t)} h_{t}^{i}}$, ensuring that $\Bar{h}_t^{n} \in \left[0, 1\right]$. Algorithm \ref{alg:alg2-expert-selection} summarizes the SAMOCP method. {It can be observed from  \eqref{eq:lifetime} that the maximum number of active experts (MOCP instances) at each time $t$ is \(g\lfloor \log_2 t \rfloor\). Hence, the complexity of SAMOCP is of order \(\mathcal{O}(T\log_2 T)\).} \par
\begin{algorithm}
\caption{Strongly Adaptive Multi-model Ensemble Online Conformal Prediction (SAMOCP)}\label{alg:alg2-expert-selection}
\begin{algorithmic}
\REQUIRE $\alpha \in [0,1]$, hyperparameters $\eta \ge 0, \epsilon \in (0, 1)$, and $\sigma > 1$.
\FOR{$t \in [T]$}
    \STATE Create new expert \(n\) (where \(n=t\)) by Algorithm \ref{alg:alg1-model-selection}($\alpha_{t-1}^{\hat{m}{\hat{n}}}, \eta, \epsilon^n$).
    \STATE Remove experts whose lifetime has been finished.
    \STATE Every active expert selects miss coverage probability from \(M\) options.
    \STATE Calculate normalized weights by $\Bar{h}_t^{n} = \frac{h_t^{n}}{\sum_{i \in {\cal A}(t)} h_{t}^{i}}$. 
    \STATE Select one miss overage probability from active experts according to PMF \(\boldsymbol{\Bar{h}}_t = (\Bar{h}_{t}^n)_{n \in {\cal A}(t)}\).
    \STATE Construct prediction set for \(X_t\) using selected miss coverage probability.
    \STATE Observe true label \(Y_t^{true}\).
    \FOR{\(n \in {\cal A}(t)\)}
        \STATE Obtain learner loss \(L(\Bar{\alpha}_t^{{\hat{m}}{{\hat{n}}}}, \alpha_t^{\hat{m}{\hat{n}}})\).
        \STATE Update every parameters assigned to each model for expert $n$ via Algorithm \ref{alg:alg1-model-selection}.
        \STATE Update \(h_{t+1}^{n}\) with \eqref{eq:updateh}.
    \ENDFOR
\ENDFOR
\end{algorithmic}
\end{algorithm}
Let $CovE(T) := \left|\frac{1}{T} \sum_{t=1}^{T} \mathbb{E}[err_t] - \alpha\right|$ represent the coverage error. In a dynamic setting where multiple experts are incorporated, each including $M$ miss coverage probabilities,  the expected error is calculated as \(\mathbb{E}[err_t] = \sum_{n=1}^{t} \sum_{m=1}^{M} \Bar{h}_t^{n} \Bar{w}_t^{mn} err_t^{mn}\), where \(mn\) represents the $m$th model by expert $n$. Using the two following theorems, we prove that SAMOCP has bounded coverage error and achieves strongly adaptive regret across any time interval of arbitrary width (Proofs can be found \ref{pr:samocp-coverage-proof} and \ref{pr:sregret-samocp-proof}).
\begin{theorem}\label{thm:coverage-alg2}
For any \(T \geq 1\) and any \(\gamma \in \left(\frac{1}{2}, 1\right)\), Algorithm \ref{alg:alg2-expert-selection} achieves the coverage error bound
\begin{equation}
CovE(T) \leq \mathcal{O}\left(\inf_{\gamma} \left\{T^{\frac{1}{2}-\gamma} + T^{\gamma-1} \beta_{\gamma}(T)\right\}\right), 
\label{eq:cove-samocp}
\end{equation}
where \(\beta_{\gamma}(T)\) measures the smoothness of model weights within experts and the cumulative gradient norm for each model within experts. The definition of \(\beta_{\gamma}(T)\) is provided in detail in equation \eqref{eq:symbol-for-coveragetheorem-alg2} in the Appendix \ref{pr:samocp-coverage-proof}. If there exists a \(\gamma \in \left(\frac{1}{2}, 1\right)\) such that \(\beta_{\gamma}(T) \leq \Tilde{\mathcal{O}}(T^{\theta})\) where \(\theta < 1 - \gamma\), then the coverage bound \eqref{eq:cove-samocp} will be \(CovE(T) \leq \Tilde{\mathcal{O}} \left(T^{-\min\left(\frac{1}{2}-\gamma, \gamma - 1 + \theta\right)}\right) = \mathbf{o}_T(1).\)
\end{theorem}

\begin{theorem}
\label{thm:regret-alg2}
Algorithm \ref{alg:alg2-expert-selection} achieves strongly adaptive regret over any interval $I \subseteq [T]$ and positive constants A, B, as follows
\begin{align}
    &\sum_{t \in I} \sum_{n \in {\cal A}(t)} \sum_{m=1}^M \Bar{h}_t^{n} \Bar{w}_t^{mn} L(\Bar{\alpha}_t^{mn}, \alpha_{t}^{mn}) - \sum_{t \in I} L(\Bar{\alpha}_t^{m^*{n^*}}, \alpha^{m^*n^*}) \leq  A\sqrt{|I|} + B\ln{T} \sqrt{|I|},
    \label{eq:regretbound-samocp}
\end{align}
where
\begin{equation}
\alpha^{m^*n^*} = \argmin_{\alpha^{m^*n}} \sum_{t \in I} L(\Bar{\alpha}_t^{m^*n}, \alpha^{m^*n}).
  \label{eq:optimal-alpha-static-samocp}
\end{equation}
Note that in equation \eqref{eq:optimal-alpha-static-samocp}, $\alpha^{m^*n}$ represents the miss coverage probability assigned to the best model for expert  \(n\), as obtained by equation \eqref{eq:regret-mocp-optimals}.
\end{theorem}
The miss coverage probability \(\alpha^{m^*n^*}\) in \eqref{eq:optimal-alpha-static-samocp},  is related to specific interval \(I\), which can vary across different intervals with distinct distributions in a dynamic environment.  In such settings, there is no fixed miss coverage probability \(\alpha^{m^*n^*}\)   that can be consistently applied over various time intervals. This necessitates establishing that SAMOCP has bounded regret in dynamic environments with respect to the time-varying benchmark in \eqref{eq:best-miss-dynamic}, as demonstrated by the following lemma. The proof is provided in \ref{pr:dregret-samocp-proof}.
\begin{lemma} 
\label{lem:dynamic-regret-samocp}
By defining the variation of the loss function to be
\begin{equation}
V(L(\cdot)_{t=1}^T)  := \sum_{t=1}^{T} \max_{\{m \in[M], n \in {\cal A}(t)\}} \left| L(\Bar{\alpha}_{t+1}^{mn}, \alpha_{t+1}^{mn}) - L(\Bar{\alpha}_{t}^{mn}, \alpha_{t}^{mn}) \right|.
    \label{eq:variability-env}
\end{equation}
We establish the following bound for the dynamic regret of Algorithm~\ref{alg:alg2-expert-selection}
\begin{equation}
    \sum_{t=1}^T \sum_{n \in {\cal A}(t)} \sum_{m=1}^M \Bar{h}_t^{n} \Bar{w}_t^{mn} L(\Bar{\alpha}_t^{mn}, \alpha_{t}^{mn}) - \sum_{t=1}^T L(\Bar{\alpha}_t^{m^*{n^*}}, \alpha_t^{m^*{n^*}}) \leq \Tilde{\mathcal{O}}(T^{\frac{2}{3}} V^{\frac{1}{3}}(L(.)_{t=1}^T) )
    \label{eq:dynamic-reg}
\end{equation}
where \(\Tilde{\mathcal{O}}\) suppresses positive constants and polylogarithmic factors, e.g., \(\log T\).  Also the best miss coverage probability at each time \(t\) can be obtained by
\begin{equation}
    \alpha_t^{m^*n^*}   = \argmin_{\{m \in [M], n \in {\cal A}(t)\}} L(\Bar{\alpha}_t^{mn}, \alpha_t^{mn}).
    \label{eq:best-miss-dynamic}
\end{equation}
\end{lemma}
Lemma \ref{lem:dynamic-regret-samocp} establishes that the dynamic regret of SAMOCP \eqref{eq:dynamic-reg} depends on the variation of the loss functions \eqref{eq:variability-env}. In addition, it can be obtained from \eqref{eq:dynamic-reg} that SAMCOP achieves sublinear regret if the variation of the loss function is also sublinear, i.e., \( V(L(.)_{t=1}^T) = \mathbf{o}(T)\).
\section{Experiments}
\label{sec:experiments}
In this section, the performance of the proposed method, SAMOCP, is assessed within the context of classification tasks. We conduct a comprehensive comparison with recently proposed methods in online conformal prediction for dynamic environments within classification tasks. The section begins with a detailed explanation of the experimental settings, followed by a discussion of the results. Note that throughout the experiments in this section, the desired miss coverage probability \(\alpha\) is $0.1$. All experiments were performed on a workstation with NVIDIA RTX A4000 GPU. Codes are available at hyperref{https://github.com/erfanhajihashemi/Multi-model-Ensemble-Conformal-Prediction-in-Dynamic-Environments}.\par
\textbf{Dataset:} We utilize corrupted versions of CIFAR-10 and CIFAR-100 \citep{krizhevsky2009}, known as CIFAR-10C and CIFAR-100C \citep{hendrycks2019}. These datasets consist of $15$ generated corruptions spanning 5 distinct levels of severity. The evaluation encompasses two settings: sudden and gradual distribution shifts. For both settings, the data sequence is split into batches of $500$ data samples each. The severity of corruption changes (increases or decreases) after each batch of data.  In the sudden shifts, the severity level alternates between the version of the data without any corruption (severity level $0$) and the most severe corruption (severity level $5$). In the gradual setting, severity starts at level $0$ and increases one by one after each batch until it reaches level $5$. After reaching level $5$, the severity decreases one by one and goes back to level $0$ in subsequent batches. This cycle of increasing and decreasing severity continues throughout the experiment. Also, additional experiments on TinyImageNet-C \citep{hendrycks2019} and synthetic data are provided in the Appendix, Section \ref{sec:add-experimment}. \par
\textbf{Baselines and experimental settings:} We employ ResNet-50, ResNet-18 \citep{he2016}, GoogLeNet \citep{szegedy2015}, and DenseNet-121 \citep{huang2017} as candidate learning models. Each active expert consists of all these learning models and needs to select the appropriate model during its active interval. The proposed method is compared with the most recent adaptive conformal prediction algorithms designed for dynamic environments, including FACI \citep{gibbs2022}, ScaleFreeOGD \citep{bhatnagar2023}, and SAOCP \citep{bhatnagar2023}. FACI employs a fixed number of active experts over all time steps, with each expert assigned one of the candidate learning rates for updating the miss coverage probability. ScaleFreeOGD reduces the learning rate based on the cumulative norms of gradients \citep{orabona2018}. SAOCP allows each expert to have its own active interval, within which it operates similarly to ScaleFreeOGD. In order to show how SAMOCP results in more efficient sets compared to SAOCP in a multi-model setting, we developed a multi-model ensemble version of SAOCP, denoted as SAOCP(MM), where each expert consists of \(M\) update rules, each corresponding to a different learning model. This approach follows our multi-model approach but employs a similar rule to SAOCP for updating weights. To determine the value of \(g\), we employed a grid search approach within the candidates \(\{4,8,16,24,32,48,64\}\). The one that led to the smallest prediction set size (Avg Width) while maintaining reasonable coverage and runtime was selected, which was \(g=8\). While the hyperparameter \(g\) is set to $8$ for both SAMOCP and SAOCP(MM), it is set to $32$ for SAOCP, as in \citep{bhatnagar2023}. Since $4$ learning models are incorporated in this section, the maximum number of updates  at each time \(t\) in SAMOCP, \(Mg\lfloor \log_2 t \rfloor\), is equal to that in SAOCP and SAOCP(MM), which is \(32\lfloor \log_2 t \rfloor\). Meanwhile, note that randomness might be undesirable in practice, 
 the predicted miss coverage in SAMOCP is calculated in a deterministic fashion, i.e., \( \alpha_t =  \sum_{n \in {\cal A}(t)} \sum_{m=1}^M \Bar{h}_t^{n} \Bar{w}_t^{mn} \alpha_{t}^{mn} \). For every experiment conducted on the synthetic dataset, CIFAR-10C, CIFAR-100C, parameters $\epsilon$, $\sigma$, and $\eta$ were selected through grid search, with values of $0.9$, $140$, and $0.05$, respectively. \par
\textbf{Score Functions:} We utilized the nonconformity score defined as in \citep{angelopoulos2020} to construct prediction sets. Let
\begin{equation}
   S^{m}(X,Y) = \xi \sqrt{\max([k_Y - k_{reg}], 0)} + U_t\hat{f}_Y^{m}(X) + \rho (X,Y),
  \label{eq:nonconformity score}   
\end{equation}
where \( \hat{f}_Y^{m}(X) \) denotes the probability of predicting label \( Y \) for input \( X \) by model \( m \), and \( U_t \) is a random variable sampled from a uniform distribution over the interval \([0, 1]\). The term \( k_Y:= |\{Y' \in \mathcal{Y} \mid  \hat{f}_{Y'}^{m}(X) \geq \hat{f}_Y^{m}(X) \}|\) denotes the number of labels that have a higher or equal predicted probability than label \( Y \) according to the model’s output probability distribution, e.g., the softmax output. \( \rho (X, Y):= \sum_{Y'=1}^{K} \hat{f}_{Y'}^{m}(X) \mathbb{I}[\hat{f}_{Y'}^{m}(X) > \hat{f}_Y^{m}(X)] \) sums up the probabilities of all labels that have a higher predicted probability than label \( Y \). The hyperparameters \( \xi \) and \( k_{reg} \) are set to $0.02$ and $5$ for CIFAR-100C, and $0.1$ and $1$ for Cifar-10C, respectively.\par
\textbf{Evaluation Metrics:} Coverage measures the percentage of instances where the true label is included in the prediction sets outputted by the conformal prediction algorithm over the period $[T]$. Avg Width refers to the average size of these prediction sets. Adaptive regret is calculated for time intervals of length 100. The metric Avg Regret represents the average of these regret values across the entire time horizon $[T]$. Run Time indicates the time required to complete each iteration of the algorithm. Lastly, Single Width measures the probability that prediction sets contain exactly one element while accurately covering the true label, highlighting cases that are most informative for predictions.
\begin{table*}[ht]
\centering
\caption{Results on the CIFAR-100C dataset with a gradual distribution shift. The target coverage is $90\%$, and the average regret is calculated over an interval size of $100$. Bold numbers denote the best results in each column. SAMOCP achieves the best performance in terms of average width, average regret, and single width.}
\label{tab:cifar-100-gradual}
\scriptsize 
\setlength{\tabcolsep}{5pt} 
\begin{tabular}{@{}l >{\raggedright\arraybackslash}p{2cm} cccc c@{}}
\toprule
Model & Method & Coverage (\%) & Avg Width & Avg Regret($\times 10^{-3}$) & Run Time & Single Width \\
\midrule
 & SAMOCP & 88.16 $\pm$ 0.18 & \textbf{5.43 $\pm$ 0.28} & \textbf{0.92 $\pm$ 0.07} & 34.87 $\pm$ 0.67 & \textbf{0.29 $\pm$ 0.01} \\
 & SAOCP(MM) & 85.89 $\pm$ 0.84 & 6.61 $\pm$ 0.26 & 4.42 $\pm$ 0.51 & 47.80 $\pm$ 0.22 & 0.27 $\pm$ 0.01 \\ \\
 DenseNet-121 & FACI & 89.64 $\pm$ 0.28 & 5.77 $\pm$ 0.62 & 1.18 $\pm$ 0.74 & 8.22 $\pm$ 0.07 & 0.28 $\pm$ 0.01 \\
 & ScaleFreeOGD & \textbf{89.97 $\pm$ 0.02} & 6.04 $\pm$ 0.10 & 1.55 $\pm$ 0.06 & \textbf{3.02 $\pm$ 0.06} & 0.26 $\pm$ 0.01 \\
 & SAOCP & 88.90 $\pm$ 0.10 & 5.80 $\pm$ 0.08 & 2.04 $\pm$ 0.04 & 40.74 $\pm$ 0.31 & 0.27 $\pm$ 0.00 \\ \\
ResNet-50 & FACI & 89.67 $\pm$ 0.33 & 6.50 $\pm$ 0.57 & 1.24 $\pm$ 0.90 & 8.45 $\pm$ 0.09 & 0.26 $\pm$ 0.01 \\
 & ScaleFreeOGD & \textbf{89.97 $\pm$ 0.02} & 6.72 $\pm$ 0.13 & 1.58 $\pm$ 0.07 & 3.12 $\pm$ 0.04 & 0.25 $\pm$ 0.01 \\
 & SAOCP & 88.79 $\pm$ 0.14 & 6.48 $\pm$ 0.13 & 2.08 $\pm$ 0.10 & 41.35 $\pm$ 0.34 & 0.25 $\pm$ 0.00 \\ \\
ResNet-18 & FACI & 89.56 $\pm$ 0.28 & 6.82 $\pm$ 0.77 & 1.17 $\pm$ 0.75 & 8.39 $\pm$ 0.06 & 0.25 $\pm$ 0.01 \\
 & ScaleFreeOGD & 89.96 $\pm$ 0.02 & 7.29 $\pm$ 0.19 & 1.55 $\pm$ 0.07 & 3.05 $\pm$ 0.04 & 0.23 $\pm$ 0.01 \\
 & SAOCP & 88.76 $\pm$ 0.22 & 6.9 $\pm$ 0.17 & 2.06 $\pm$ 0.07 & 41.23 $\pm$ 0.26 & 0.24 $\pm$ 0.01 \\ \\
GoogLeNet & FACI & 89.63 $\pm$ 0.30 & 6.33 $\pm$ 0.74 & 1.10 $\pm$ 0.78 & 8.30 $\pm$ 0.09 & 0.27 $\pm$ 0.01 \\
 & ScaleFreeOGD & 89.96 $\pm$ 0.01 & 6.71 $\pm$ 0.15 & 1.52 $\pm$ 0.06 & 3.04 $\pm$ 0.04 & 0.24 $\pm$ 0.00 \\
 & SAOCP & 88.68 $\pm$ 0.11 & 6.38 $\pm$ 0.12 & 2.07 $\pm$ 0.07 & 41.13 $\pm$ 0.39 & 0.26 $\pm$ 0.01 \\
\bottomrule
\end{tabular}
\end{table*}
\subsection{Results}
Table \ref{tab:cifar-100-gradual} presents the performance of SAMOCP for the classification task on the CIFAR-100C dataset under a gradual shift setting, where each method receives $8,550$ data points sequentially. It is evident that the performance of previous methods, particularly in terms of prediction set size and single-width prediction sets, depends on the learning model employed. The proposed method SAMOCP outperforms existing methods by creating smaller prediction sets, lower regret, and more single-width prediction sets that correctly cover the true label, while also achieving coverage close to the targeted level. It is noteworthy that SAMOCP surpasses every variant of previous methods in these aspects. Furthermore, SAMOCP is faster than SAOCP and SAOCP(MM), despite having the same maximum number of updates at each time \(t\). \par
In dynamic environments, data distribution does not necessarily shift gradually. Instead, we may encounter abrupt distribution shifts, with significant differences between data distributions in two successive time slots. To demonstrate how SAMOCP behaves in such environments, another experiment was conducted, in which SAMOCP can successfully track these sharp transitions and select a suitable learning model for creating a prediction set. Experimental results on CIFAR-10C are detailed in Table \ref{tab:cifar-10-sudden}, where the proposed algorithm again outperforms previous methods in terms of prediction set size, regret, and single width prediction sets that accurately cover the true labels while maintaining coverage close to the target value. \par
\begin{figure}
    \centering
    \begin{subfigure}[b]{0.52\textwidth}
        \includegraphics[width=\textwidth]{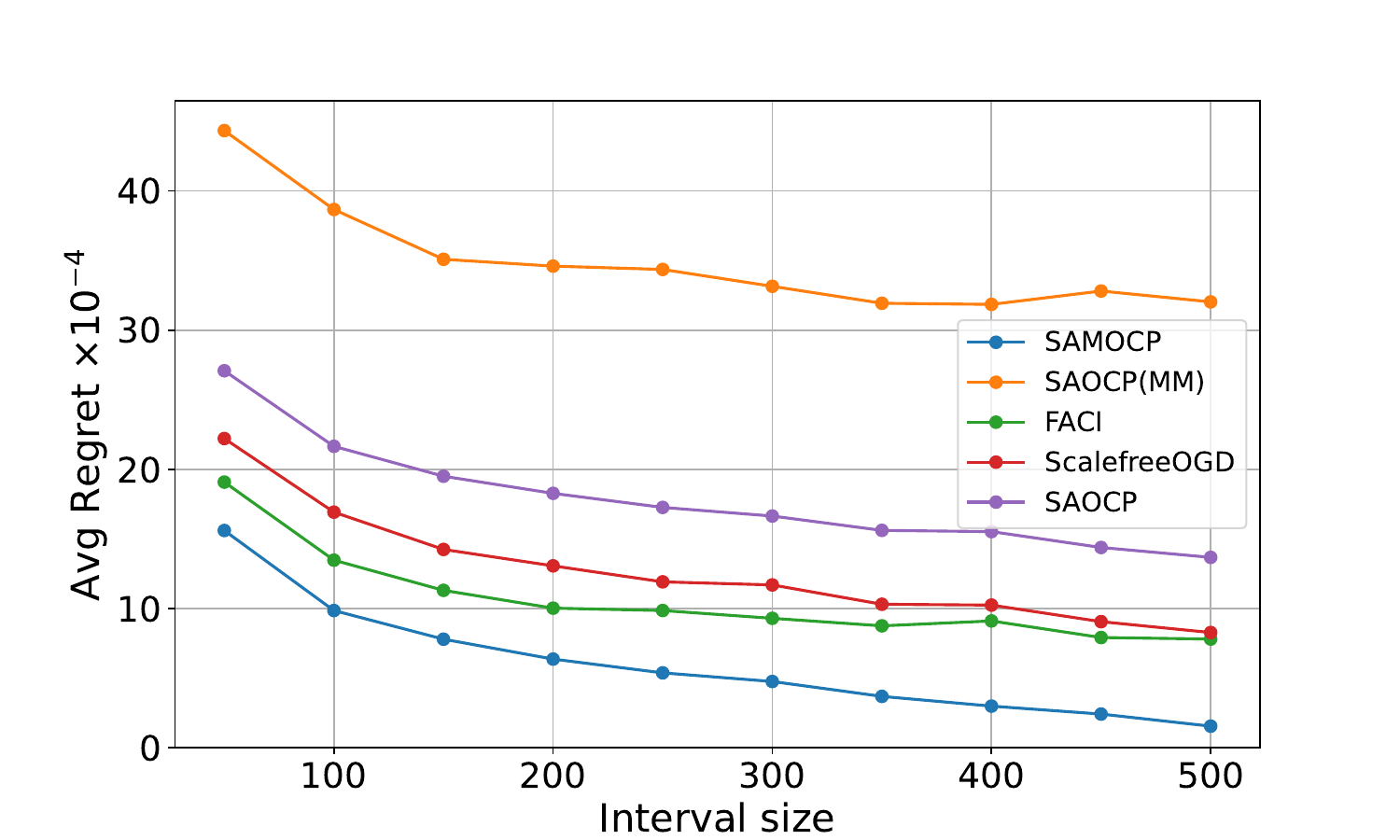}
        \caption{CIFAR-10C}
    \end{subfigure}
    \hfill\hspace{-10mm}
    \begin{subfigure}[b]{0.52\textwidth}
        \includegraphics[width=\textwidth]{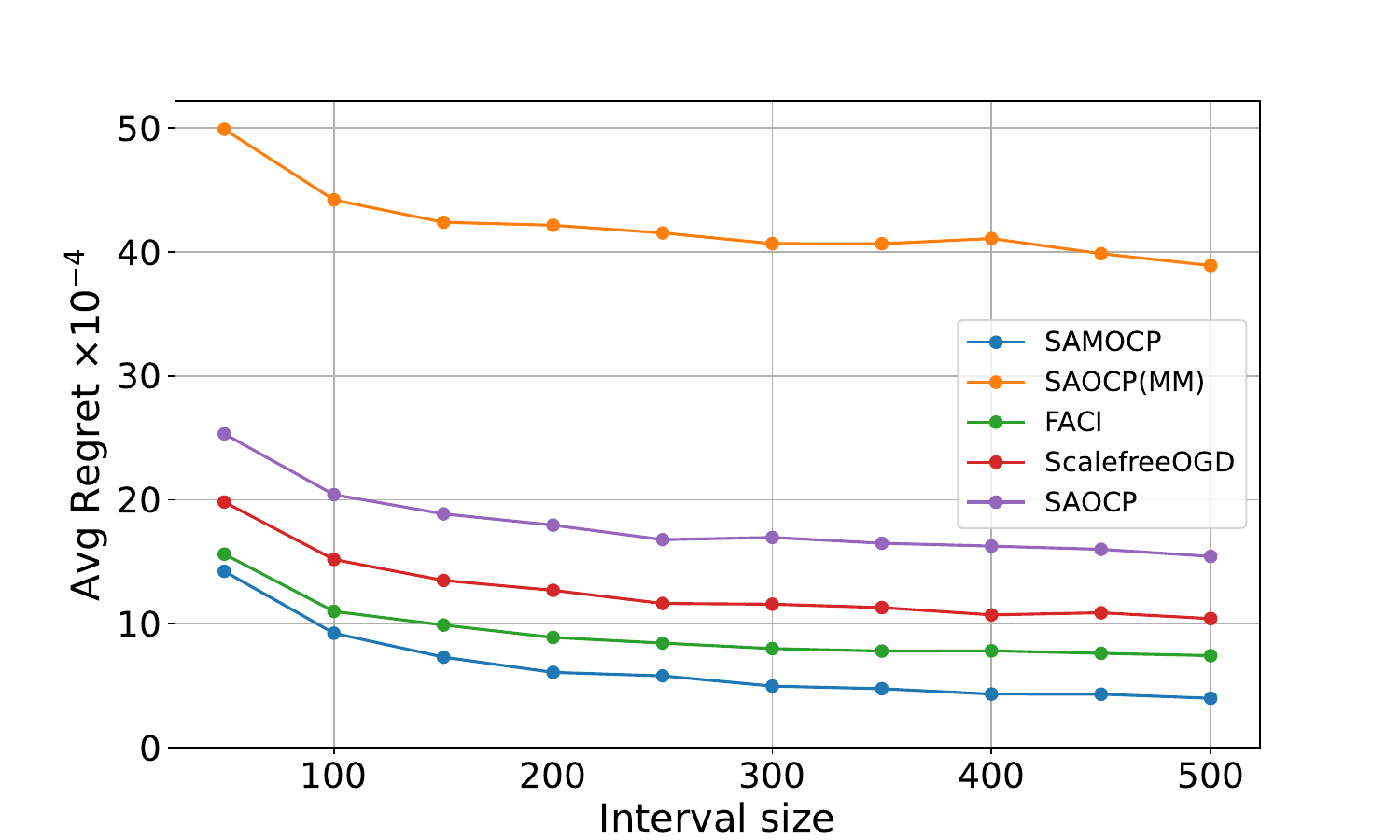}
        \caption{CIFAR-100C}
    \end{subfigure}
    \caption{Evaluation of average regret over different interval sizes (\(50, 100, \ldots, 500\)). Note that for previous methods relying on a single model, the lowest regret across the $4$ learning models is selected.}

    \label{fig:regret-figure}
\end{figure}

To demonstrate that SAMOCP achieves the lowest regret over different intervals compared to existing methods, we illustrate the regret for various interval sizes in Figure \ref{fig:regret-figure}. For each existing method, there are $4$ different regrets corresponding to the $4$ learning models used and the lowest regret is depicted. The results show that our method consistently leads to lower regret than the best version of each previous method across different learning models. Note that a lower regret implies that the algorithm adapts faster to changes. The regret calculated over different time intervals indicates the algorithm’s adaptivity in capturing the distribution shift at different time scales. Therefore, Figure \ref{fig:regret-figure} indicates that the SAMOCP can adapt faster to distribution shifts compared to benchmarks in various time scales. Furthermore, we include experiments on synthetic data and another real dataset, TinyImageNet-C, using different sets of learning models that do not necessarily contain $4$ models in the Appendix, Section \ref{sec:add-experimment}. This demonstrates how SAMOCP can rely on a mixture of learning models over the period $[T]$ and select the appropriate one for each distribution setting.
\begin{table*}[ht]
\centering
\caption{Results on the CIFAR-10C dataset with a sudden distribution shift. The target coverage is \(90\%\), and the average regret is calculated over an interval size of $100$. Bold numbers denote the best results in each column. SAMOCP achieves the best performance in terms of average width, average regret, and single width.}
\label{tab:cifar-10-sudden}
\scriptsize 
\setlength{\tabcolsep}{5pt} 
\begin{tabular}{@{}l >{\raggedright\arraybackslash}p{2cm} cccc c@{}}
\toprule
Model & Method & Coverage (\%) & Avg Width & Avg Regret($\times 10^{-3}$) & Run Time & Single Width \\
\midrule
 & SAMOCP & 88.37 $\pm$ 0.23 & \textbf{1.24 $\pm$ 0.06} & \textbf{0.98 $\pm$ 0.11} & 33.75 $\pm$ 0.34 & \textbf{0.69 $\pm$ 0.03} \\
 & SAOCP(MM) & 86.80 $\pm$ 2.39 & 1.45 $\pm$ 0.13 & 3.87 $\pm$ 1.05 & 47.08 $\pm$ 0.19 & 0.56 $\pm$ 0.05 \\ \\
 DenseNet-121 & FACI & 89.57 $\pm$ 0.37 & 1.30 $\pm$ 0.12 & 1.46 $\pm$ 0.73 & 8.11 $\pm$ 0.10 & 0.68 $\pm$ 0.05 \\
 & ScaleFreeOGD & \textbf{89.99 $\pm$ 0.01} & 1.46 $\pm$ 0.02 & 1.71 $\pm$ 0.04 & 2.92 $\pm$ 0.07 & 0.52 $\pm$ 0.02 \\
 & SAOCP & 88.77 $\pm$ 0.18 & 1.41 $\pm$ 0.02 & 2.24 $\pm$ 0.06 & 39.62 $\pm$ 0.22 & 0.54 $\pm$ 0.01 \\ \\
ResNet-50 & FACI & 89.74 $\pm$ 0.35 & 1.50 $\pm$ 0.04 & 1.35 $\pm$ 0.93 & 8.11 $\pm$ 0.08 & 0.55 $\pm$ 0.01 \\
 & ScaleFreeOGD & 89.98 $\pm$ 0.01 & 1.52 $\pm$ 0.01 & 1.71 $\pm$ 0.05 & \textbf{2.89 $\pm$ 0.04} & 0.54 $\pm$ 0.01 \\
 & SAOCP & 89.12 $\pm$ 0.08 & 1.51 $\pm$ 0.01 & 2.17 $\pm$ 0.06 & 40.25 $\pm$ 0.27 & 0.53 $\pm$ 0.01 \\ \\
ResNet-18 & FACI & 89.63 $\pm$ 0.34 & 1.36 $\pm$ 0.13 & 1.52 $\pm$ 0.76 & 8.11 $\pm$ 0.08 & 0.64 $\pm$ 0.05 \\
 & ScaleFreeOGD & \textbf{89.99 $\pm$ 0.01} & 1.52 $\pm$ 0.02 & 1.69 $\pm$ 0.06 & 2.91 $\pm$ 0.06 & 0.49 $\pm$ 0.01 \\
 & SAOCP & 88.83 $\pm$ 0.06 & 1.48 $\pm$ 0.02 & 2.24 $\pm$ 0.07 & 40.16 $\pm$ 0.22 & 0.51 $\pm$ 0.01 \\ \\
GoogLeNet & FACI & 89.73 $\pm$ 0.34 & 1.43 $\pm$ 0.06 & 1.41 $\pm$ 0.89 & 8.10 $\pm$ 0.10 & 0.58 $\pm$ 0.02 \\
 & ScaleFreeOGD & 89.99 $\pm$ 0.02 & 1.46 $\pm$ 0.01 & 1.70 $\pm$ 0.06 & \textbf{2.89 $\pm$ 0.04} & 0.55 $\pm$ 0.00 \\
 & SAOCP & 89.09 $\pm$ 0.14 & 1.44 $\pm$ 0.01 & 2.17 $\pm$ 0.08 & 40.13 $\pm$ 0.17 & 0.55 $\pm$ 0.01 \\
\bottomrule
\end{tabular}
\end{table*}

\section*{Acknowledgement}
Work in this paper is supported by NSF ECCS 2207457 and NSF ECCS 2412484.

\section*{Conclusion} In this study, we introduced a novel conformal prediction algorithm designed for online environments undergoing distribution shifts. Recognizing that the selection of baseline models affects the efficiency of conformal prediction, our algorithm incorporates multiple models simultaneously. For each expert, the contribution of each model is dynamically adjusted based on its time-evolving weight. We demonstrated that our proposed method SAMOCP achieves strongly adaptive regret across any time interval of arbitrary width and maintains valid coverage. Experimental results in environments with both gradual and sudden distribution shifts indicated that our algorithm produces more informative prediction sets and achieves a coverage rate close to the target value, compared to those created by previous methods using their best baseline models.
\bibliographystyle{plainnat}
\bibliography{references}
\newpage
\appendix
\section{Proofs}
\label{sec:proof}
\begin{lemma}\label{lem:bounded-alpha}
For every \(\alpha \in [0,1]\) and learning rate \(\eta > 0\), adaptive miss coverage probability  \(\alpha_t^{m}\) for any model $m \in [M]$ and \(t \in [T]\) is bounded as
\[
\alpha_t^{m} \in \left[-\eta, 1+\eta \right].
\]
\end{lemma}
\subsection{Proof of lemma \ref{lem:bounded-alpha}}
\label{pr:boundeda-alpha-proof}
Based on equation \eqref{eq:update-rule}, we have
\begin{equation}
  |\alpha_{t}^{m} - \alpha_{t-1}^{m}| = \eta \left| \frac{(\alpha - err_{t-1}^{m})}{\sqrt{\sum_{\tau=1}^{t-1} \|\nabla_{\alpha_{\tau}^{m}} L(\Bar{\alpha}_{\tau}^m, \alpha_{\tau}^{m})\|_2^2 }} \right| \leq \eta.
  \label{eq:update-rule-in-lemma}
\end{equation}
We prove this lemma using a contradiction. Suppose there exists a $\hat{t}$ such that $\alpha_{\hat{t}}^{m} \notin [-\eta, 1+\eta]$, where $\hat{t} \geq 2$ is the smallest such time index. We first prove the case of violating the upper bound; by contradiction, assume $\alpha_{\hat{t}}^{m} > 1 + \eta$. According to equation \eqref{eq:update-rule-in-lemma}, this would necessitate that $\alpha_{\hat{t}-1}^{m} \geq 1$. Given that we assumed $\hat{t}$ is the smallest time index to violate the upper bound, it should follow that $\alpha_{\hat{t}-1}^{m} \leq 1+\eta$. However, $\alpha_{\hat{t}-1}^{m} > 1 > \Bar{\alpha}_{\hat{t}-1}^m$ implies $err_{\hat{t}-1}^{m}=1$. By equation \eqref{eq:update-rule} we have
\[
\alpha_{\hat{t}}^{m} = \alpha_{\hat{t}-1}^{m} + \eta \frac{\alpha - 1}{\sqrt{\sum_{\tau=1}^{\hat{t}-1} \|\nabla_{\alpha_{\tau}^{m}} L(\Bar{\alpha}_{\tau}^m, \alpha_{\tau}^{m})\|_2^2 }} \leq \alpha_{\hat{t}-1}^{m} \leq 1+\eta,
\]
which contradicts our assumption that $\alpha_{\hat{t}-1}^{m} > 1 + \eta$.
Next, assume $\alpha_{\hat{t}}^{m} < -\eta$. By equation \eqref{eq:update-rule-in-lemma}, we have $\alpha_{\hat{t}-1}^{m} < 0$. Given that $\hat{t}$ is the smallest index that violates the lower bound of the lemma, it must hold that $\alpha_{\hat{t}-1}^{m} \geq -\eta$. Considering $\alpha_{\hat{t}-1}^{m} < 0 < \Bar{\alpha}_{\hat{t}-1}^m$, we deduce that $err_{\hat{t}-1}^{m} = 0$. Therefore, by equation \eqref{eq:update-rule}, we have
\[
\alpha_{\hat{t}}^{m} = \alpha_{\hat{t}-1}^{m} + \eta \frac{\alpha}{\sqrt{\sum_{\tau=1}^{\hat{t}-1} \|\nabla_{\alpha_{\tau}^{m}} L(\Bar{\alpha}_{\tau}^m, \alpha_{\tau}^{m})\|_2^2 }} \geq \alpha_{\hat{t}-1}^{m} \geq -\eta,
\]
which contradicts our initial assumption that $\alpha_{\hat{t}}^{m} < -\eta$.
\subsection{Proof of Theorem \ref{thm:regret-alg1}, Regret for MOCP:}
\label{pr:regret-MOCP-proof}
 Algorithm \ref{alg:alg1-model-selection} has the following regret bound
\[
\sum_{t=1}^{T} \sum_{m=1}^{M} \Bar{w}_t^{m}L(\Bar{\alpha}_t^m, \alpha_t^{m}) -  \sum_{t=1}^{T} L(\Bar{\alpha}_t^{m^*}, \alpha^{m^*}) \le \sqrt{T}\left(\frac{(1+2\eta)^2}{2\eta} + \frac{\eta}{2\alpha} + \ln{M} + (1+\eta)^2\right), \]
Where $\alpha^{m^*}$ can be obtained via \eqref{eq:regret-mocp-optimals}. To prove the theorem, we first introduce and prove the following two lemmas
\begin{lemma}\label{lem:regret-bound-lem1} for miss coverage probability  assigned to any model $\Tilde{m} \in [M]$, we have the following bound
\[
\sum_{t=1}^{T} L(\Bar{\alpha}_t^{\Tilde{m}}, \alpha_t^{\Tilde{m}}) -
\sum_{t=1}^{T} L(\Bar{\alpha}_t^{\Tilde{m}}, \alpha^{\Tilde{m}}) \leq
\frac{\sqrt{T}}{2\eta}(1+2\eta)^2 + \frac{\eta \sqrt{T}}{2\alpha},
\]
where $\alpha^{\Tilde{m}} = \argmin_{\alpha_t^{\Tilde{m}}} \sum_{t=1}^{T} L(\Bar{\alpha}_t^{\Tilde{m}}, \alpha_t^{\Tilde{m}})$. \\
\textbf{Proof:} We first begin with
\[
(\alpha_{t+1}^{\Tilde{m}} - \alpha^{\Tilde{m}})^2 = 
(\alpha_{t}^{\Tilde{m}} - \eta \frac{\nabla_{\alpha_t^{\Tilde{m}}} L(\Bar{\alpha}_t^{\Tilde{m}}, \alpha_t^{\Tilde{m}})}{\sqrt{ \sum_{\tau=1}^{t} \|\nabla_{\alpha_{\tau}^{\Tilde{m}}} L(\Bar{\alpha}_{\tau}^{\Tilde{m}}, \alpha_{\tau}^{\Tilde{m}})\|_2^2 }} - \alpha^{\Tilde{m}})^2.
\]
Then define adaptive learning rate \(\eta_t\) \citep{duchi2011adaptive, hazan2007adaptive} as \[\eta_t := \frac{\eta}{\sqrt{ \sum_{\tau=1}^{t} \|\nabla_{\alpha_{\tau}^{\Tilde{m}}} L(\Bar{\alpha}_{\tau}^{\Tilde{m}}, \alpha_{\tau}^{\Tilde{m}})\|_2^2 }}. \] 
So we have
\[
(\alpha_{t+1}^{\Tilde{m}} - \alpha^{\Tilde{m}})^2 = 
(\eta_t \nabla_{\alpha_t^{\Tilde{m}}} L(\Tilde{\alpha}_t^{\Tilde{m}},\alpha_t^{\Tilde{m}}))^2 + 
(\alpha_{t}^{\Tilde{m}} - \alpha^{\Tilde{m}})^2 - 
2\eta_t (\alpha_{t}^{\Tilde{m}} - \alpha^{\Tilde{m}}) 
\nabla_{\alpha_t^{\Tilde{m}}} L(\Tilde{\alpha}_t^{\Tilde{m}}, \alpha_t^{\Tilde{m}}).
\]
Therefore,
\[
 (\alpha_{t}^{\Tilde{m}} - \alpha^{\Tilde{m}}) 
\nabla_{\alpha_t^{\Tilde{m}}} L(\Bar{\alpha}_t^{\Tilde{m}}, \alpha_t^{\Tilde{m}}) = 
\frac{(\alpha_{t}^{\Tilde{m}} - \alpha^{\Tilde{m}})^2 - (\alpha_{t+1}^{\Tilde{m}} - \alpha^{\Tilde{m}})^2 }{2\eta_t} + \frac{\eta_t}{2} (\nabla_{\alpha_t^{\Tilde{m}}} L(\Bar{\alpha}_t^{\Tilde{m}}, \alpha_t^{\Tilde{m}}))^2.
\]
Since the loss function \eqref{eq:loss-model-m} is convex, we have the following inequality
\begin{equation}
  L(\Bar{\alpha}_t^{\Tilde{m}}, \alpha_t^{\Tilde{m}}) -
 L(\Bar{\alpha}_t^{\Tilde{m}}, \alpha^{\Tilde{m}}) \leq
(\alpha_t^{\Tilde{m}} - \alpha^{\Tilde{m}}) \nabla_{\alpha_t^{\Tilde{m}}} L(\Bar{\alpha}_t^{\Tilde{m}}, \alpha_t^{\Tilde{m}}).   
\label{eq:cvx-loss}
\end{equation}
By summing \eqref{eq:cvx-loss} over $t \in [T]$ we have
\begin{align}
    & \sum_{t=1}^T \left(L\left(\Bar{\alpha}_t^{\Tilde{m}}, \alpha_t^{\Tilde{m}}\right) -L\left(\Bar{\alpha}_t^{\Tilde{m}}, \alpha^{\Tilde{m}}\right)\right) \nonumber \\
    & \leq \sum_{t=1}^T \frac{(\alpha_{t}^{\Tilde{m}} - \alpha^{\Tilde{m}})^2 - (\alpha_{t+1}^{\Tilde{m}} - \alpha^{\Tilde{m}})^2 }{2\eta_t} + \sum_{t=1}^T \frac{\eta_t}{2} \left(\nabla_{\alpha_t^{\Tilde{m}}} L(\Bar{\alpha}_t^{\Tilde{m}}, \alpha_t^{\Tilde{m}})\right)^2 \nonumber\\
    & \leq \frac{\sqrt{T}}{2\eta} \sum_{t=1}^T \left((\alpha_{t}^{\Tilde{m}} - \alpha^{\Tilde{m}})^2 - (\alpha_{t+1}^{\Tilde{m}} - \alpha^{\Tilde{m}})^2\right) + \frac{\eta}{2} \sum_{t=1}^T \frac{1}{\sqrt{ \sum_{\tau=1}^{t} \|\nabla_{\alpha_{\tau}^{\Tilde{m}}} L(\Bar{\alpha}_{\tau}^{\Tilde{m}}, \alpha_{\tau}^{\Tilde{m}})\|_2^2 }} \nonumber \\
    &  \leq \frac{\sqrt{T}}{2\eta} \left((\alpha_{1}^{\Tilde{m}} - \alpha^{\Tilde{m}})^2 - (\alpha_{T+1}^{\Tilde{m}} - \alpha^{\Tilde{m}})^2\right) + \frac{\eta}{2} \sum_{t=1}^T \frac{1}{\alpha \sqrt{T}} \overset{\text{(i)}}{\leq}
\frac{\sqrt{T}}{2\eta} (1 + 2\eta)^2 + \frac{\eta \sqrt{T}}{2\alpha},
\end{align}
where (i) used \(\alpha_t^{m} \in \left[-\eta, 1+\eta \right]\) by Lemma \ref{lem:bounded-alpha}.
\end{lemma}
\begin{lemma}\label{lem:regret-bound-lem2} For miss coverage probability assigned to any model $\Tilde{m} \in [M]$ we have the following bound
\[
\sum_{t=1}^{T} \sum_{m=1}^{M} \Bar{w}_t^{m}L(\Bar{\alpha}_t^m, \alpha_t^{m}) -
\sum_{t=1}^{T} L(\Bar{\alpha}_t^{\Tilde{m}}, \alpha_t^{\Tilde{m}}) \leq
\frac{\ln{M}}{\epsilon} + \epsilon (1+\eta)^2 T .
\]
\textbf{Proof:} Referring to the definition of $\Bar{w}_t^{m}$ in Subsection \ref{sec:model-selection} and defining $W_{t} := \sum_{m=1}^M w_{t}^{m}$, we have
\begin{align}
    & W_{T+1} = \sum_{m=1}^M w_{T+1}^{m} = \sum_{m=1}^M w_{T}^{m} \exp{\left(-\epsilon L(\Bar{\alpha}_T^m, \alpha_T^{m})\right)} = W_T \sum_{m=1}^M \Bar{w}_{T}^{m} \exp{(-\epsilon L(\Bar{\alpha}_T^m, \alpha_T^{m}))} \nonumber \\
    &\overset{\text{(i)}}{\leq} W_T \sum_{m=1}^M \Bar{w}_{T}^{m} \left(1 -\epsilon L(\Bar{\alpha}_T^m, \alpha_T^{m}) + \epsilon^2 L(\Bar{\alpha}_T^m, \alpha_T^{m})^2 \right) \nonumber \\
    & = W_T \left(1 - \epsilon \sum_{m=1}^M \Bar{w}_{T}^{m} L(\Bar{\alpha}_T^m, \alpha_T^{m}) + \epsilon^2 \sum_{m=1}^M \Bar{w}_{T}^{m}  L(\Bar{\alpha}_T^m, \alpha_T^{m})^2\right) \nonumber \\
    &\overset{\text{(ii)}}{\leq} W_T  \exp{\left(- \epsilon \sum_{m=1}^M \Bar{w}_{T}^{m} L(\Bar{\alpha}_T^m, \alpha_T^{m}) + \epsilon^2 \sum_{m=1}^M \Bar{w}_{T}^{m}  L(\Bar{\alpha}_T^m, \alpha_T^{m})^2\right)} \nonumber \\
    & \leq W_1 \exp{\left(- \epsilon \sum_{t=1}^T \sum_{m=1}^M \Bar{w}_{t}^{m} L(\Bar{\alpha}_t^m, \alpha_t^{m}) +
    \epsilon^2 \sum_{t=1}^T \sum_{m=1}^M \Bar{w}_{t}^{m}  L(\Bar{\alpha}_t^m, \alpha_t^{m})^2 \right)},
    \label{eq:lemma4-side1}
\end{align}
where $W_1 = 1$, $(\text{i})$ follows from the inequality $\exp(-\epsilon x) \leq 1 - \epsilon x + \epsilon^2 x^2$ for $|\epsilon| \leq 1$, and $(\text{ii})$ follows from $1+x \leq e^x$. On the other hand, we have
\begin{equation}
W_{T+1} \geq w_{T+1}^{\Tilde{m}} = w_1^{\Tilde{m}} \prod_{t=1}^T \exp{(-\epsilon L(\Bar{\alpha}_t^{\Tilde{m}}, \alpha_t^{\Tilde{m}}))} =
    w_1^{\Tilde{m}} \exp{(-\epsilon \sum_{t=1}^T L(\Bar{\alpha}_t^{\Tilde{m}}, \alpha_t^{\Tilde{m}}))},
    \label{eq:lemma4-side2}
\end{equation}
where $w_1^{\Bar{m}} = \frac{1}{M}$. By combining \eqref{eq:lemma4-side1} and \eqref{eq:lemma4-side2} we have
\begin{align}
    & \exp{\left(-\epsilon \sum_{t=1}^T \sum_{m=1}^M \Bar{w}_{t}^{m} L(\Bar{\alpha}_t^m, \alpha_t^{m}) + \epsilon^2 \sum_{t=1}^T \sum_{m=1}^M \Bar{w}_{t}^{m}  L(\Bar{\alpha}_t^m, \alpha_t^{m})^2\right)} \nonumber \\
    &\geq \frac{1}{M} \exp{\left(-\epsilon \sum_{t=1}^T L(\Bar{\alpha}_t^{\Tilde{m}}, \alpha_t^{\Tilde{m}})\right)}.
\end{align}
By taking the logarithm on both sides we have
\[
-\epsilon \sum_{t=1}^T \sum_{m=1}^M  \Bar{w}_{t}^{m} L(\Bar{\alpha}_t^m, \alpha_t^{m}) + 
\epsilon^2 \sum_{t=1}^T \sum_{m=1}^M \Bar{w}_{t}^{m}  L(\Bar{\alpha}_t^m, \alpha_t^{m})^2 
\geq -\ln{M} -\epsilon \sum_{t=1}^T L(\Bar{\alpha}_t^{\Tilde{m}}, \alpha_t^{\Tilde{m}}),
\]
which leads to
\begin{equation}
\sum_{t=1}^T \sum_{m=1}^M  \Bar{w}_{t}^{m} L(\Bar{\alpha}_t^m, \alpha_t^{m})
- \sum_{t=1}^T L(\Bar{\alpha}_t^{\Tilde{m}}, \alpha_t^{\Tilde{m}}) \leq 
    \frac{\ln{M}}{\epsilon} + T \epsilon (1+\eta)^2.
    \label{eq: combined-two-lemma}
\end{equation}
\end{lemma}
Now, we define the best model in the static environment as
\[
m^* = \argmin_{m \in M} \sum_{t=1}^{T} L(\Bar{\alpha}_t^{m}, \alpha_t^{m}).
\]
Then, we replace $\Tilde{m}$ with best model $m^*$ in Lemma \ref{lem:regret-bound-lem1} and \ref{lem:regret-bound-lem2}.By setting $\epsilon = \frac{1}{\sqrt{T}} $ and summing results of two lemmas we have:
\begin{align}
    &\sum_{t=1}^T \sum_{m=1}^M  \Bar{w}_{T}^{m} L(\Bar{\alpha}_t^m, \alpha_t^{m}) - \sum_{t=1}^T L(\Bar{\alpha}_t^{m^*}, \alpha_t^{m^*}) +  \sum_{t=1}^{T} L(\Bar{\alpha}_t^{m^*}, \alpha_t^{m^*}) - \sum_{t=1}^{T} L(\Bar{\alpha}_t^{m^*}, \alpha^{m^*}) \nonumber \\
    & = \sum_{t=1}^T \sum_{m=1}^M  \Bar{w}_{T}^{m} L(\Bar{\alpha}_t^m, \alpha_t^{m})
- \sum_{t=1}^{T} L(\Bar{\alpha}_t^{m^*}, \alpha^{m^*}) \leq
\sqrt{T}\left(\frac{(1+2\eta)^2}{2\eta} + \frac{\eta}{2\alpha} + \ln{M} + (1+\eta)^2\right).
\end{align}
\subsection{Proof of Theorem \ref{thm:coverage-alg2}, Coverage error for SAMOCP: }

\label{pr:samocp-coverage-proof}
 We first define expected miss coverage error as
\[\mathbb{E}[err_t] = \sum_{n=1}^{t} \sum_{m=1}^{M} \Bar{h}_t^{n} \Bar{w}_t^{mn} err_t^{mn}.\]
The proof of this theorem is based on a grouping argument. So we first divide $T$ into $\lceil T^{1 - \gamma} \rceil$ group for $\gamma \in (\frac{1}{2}, 1)$, and write the $k$th group as
\[G_k = \{t_{k-1} + 1, \ldots, \min(t_k,T)\}.\]
where $|G_k| \leq \lceil T^{\gamma} \rceil$. We also define a new variable, $H_{n:t}^{mn}$ , assigned to $m$th update rule of $n$th expert as follows
\[
H_{n:t}^{mn} := \sqrt{\sum_{\tau=n}^{t} \|\nabla_{\alpha_{\tau}^{mn}} L(\Bar{\alpha}_{\tau}^{mn}, \alpha_{\tau}^{mn}) \|_2^2}.
\]
So the update rule in \eqref{eq:update-rule} can be written as:
\begin{equation}
      \alpha_{t+1}^{mn} = \alpha_t^{mn} + \eta \frac{(\alpha - err_t^{mn})}{H_{n:t}^{mn}}.
      \label{eq:update-rule-for-thm2}
\end{equation}
For $k$th group where $2 \leq k \leq \lceil T^{1-\gamma} \rceil$, by using \eqref{eq:update-rule-for-thm2} we have
\[\mathbb{E}[err_t] - \alpha = 
 \sum_{n=1}^{t} \sum_{m=1}^{M} \Bar{h}_t^{n} \Bar{w}_t^{mn} (err_t^{mn} - \alpha) = \frac{1}{\eta}  \sum_{n=1}^{t} \sum_{m=1}^{M} \Bar{h}_t^{n} \Bar{w}_t^{mn} (\alpha_t^{mn} - \alpha_{t+1}^{mn}) H_{n:t}^{mn}.\]
Since at each time $t$ we activate an expert with lifetime as defined in \eqref{eq:lifetime}, the $n$th expert will be activated at time $t=n$. Consequently, $\Bar{h}_t^{n}$ will be $0$ for $t < n$. Therefore, we have
\begin{align}
    &\frac{1}{\eta}  \sum_{n=1}^{t} \sum_{m=1}^{M} \Bar{h}_t^{n} \Bar{w}_t^{mn} (\alpha_t^{mn} - \alpha_{t+1}^{mn}) H_{n:t}^{mn} = \frac{1}{\eta}  \sum_{n=1}^{t_{k-1}} \sum_{m=1}^{M} \Bar{h}_{t_{k-1}}^n \Bar{w}_{t_{k-1}}^{mn} (\alpha_t^{mn} - \alpha_{t+1}^{mn}) H_{n:t_{k-1}}^{mn} \nonumber \\
    &+ \frac{1}{\eta}  \sum_{n=1}^{t} \sum_{m=1}^{M} \Bar{h}_t^{n} \Bar{w}_t^{mn} (\alpha_t^{mn} - \alpha_{t+1}^{mn}) H_{n:t}^{mn} - \frac{1}{\eta}  \sum_{n=1}^{t_{k-1}} \sum_{m=1}^{M} \Bar{h}_{t_{k-1}}^n \Bar{w}_{t_{k-1}}^{mn} (\alpha_t^{mn} - \alpha_{t+1}^{mn}) H_{n:t_{k-1}}^{mn} \nonumber \\
    &= \frac{1}{\eta}  \sum_{n=1}^{t_{k-1}} \sum_{m=1}^{M} \Bar{h}_{t_{k-1}}^n \Bar{w}_{t_{k-1}}^{mn} (\alpha_t^{mn} - \alpha_{t+1}^{mn}) H_{n:t_{k-1}}^{mn} \nonumber \\
    & + \frac{1}{\eta}  \sum_{n=1}^{t} \sum_{m=1}^{M} (\Bar{h}_t^{n} \Bar{w}_t^{mn} - \Bar{h}_{t_{k-1}}^n \Bar{w}_{t_{k-1}}^{mn} \frac{H_{n:t_{k-1}}^{mn}}{H_{n:t}^{mn}}) (\alpha_t^{mn} - \alpha_{t+1}^{mn}) H_{n:t}^{mn}.
    \label{eq:inter-cov-proof-samocp}
\end{align}
Note that according to \eqref{eq:update-rule-for-thm2}, $H_{n:t}^{mn}(\alpha_t^{mn} - \alpha_{t+1}^{mn}) \leq \eta$. By summing \eqref{eq:inter-cov-proof-samocp} over $t \in G_k$ we have
\begin{align}
    &\left|\sum_{t \in G_k} (\mathbb{E}[err_t] - \alpha)\right| \leq \left|\frac{1}{\eta}  \sum_{n=1}^{t_{k-1}} \sum_{m=1}^{M} \Bar{h}_{t_{k-1}}^n \Bar{w}_{t_{k-1}}^{mn} H_{n:t_{k-1}}^{mn} \sum_{t \in G_k} (\alpha_t^{mn} - \alpha_{t+1}^{mn})\right| \nonumber \\
    &  + \left|\frac{1}{\eta} \sum_{t \in G_k}  \sum_{n=1}^{t} \sum_{m=1}^{M} (\Bar{h}_t^{n}\Bar{w}_t^{mn}  -\Bar{h}_{t_{k-1}}^n \Bar{w}_{t_{k-1}}^{mn} \frac{H_{n:t_{k-1}}^{mn}}{H_{n:t}^{mn}}) (\alpha_t^{mn} - \alpha_{t+1}^{mn}) H_{n:t}^{mn}\right| \nonumber \\
    & \leq \left| \frac{1}{\eta}  \sum_{n=1}^{t_{k-1}} \sum_{m=1}^{M} \Bar{h}_{t_{k-1}}^n \Bar{w}_{t_{k-1}}^{mn} H_{n:t_{k-1}}^{mn} (\alpha_{t_{k-1}+1}^{mn} - \alpha_{t_k+1}^{mn})\right| \nonumber \\
    &+ \left|\sum_{t \in G_k}  \sum_{n=1}^{t} \sum_{m=1}^{M} (\Bar{h}_t^{n}\Bar{w}_t^{mn} - \Bar{h}_{t_{k-1}}^n \Bar{w}_{t_{k-1}}^{mn} \frac{H_{n:t_{k-1}}^{mn}}{H_{n:t}^{mn}})  \right| \nonumber \\
    & \leq \frac{1}{\eta} \max_{\{m \in [M], n \in [t_{k-1}] \}} H_{n:t_{k-1}}^{mn} \left|\alpha_{t_{k-1}+1}^{mn} - \alpha_{t_k+1}^{mn}\right| \nonumber \\
    & + |G_k| \cdot \max_{t \in G_k }  \sum_{n=1}^{t} \sum_{m=1}^{M} \left|\Bar{h}_t^{n}\Bar{w}_t^{mn} - \Bar{h}_{t_{k-1}}^n \Bar{w}_{t_{k-1}}^{mn} \frac{H_{n:t_{k-1}}^{mn}}{H_{n:t}^{mn}}  \right| \nonumber \\
    &\leq \frac{1+ 2\eta}{\eta} \sqrt{T} + \lceil T^{\gamma} \rceil \cdot \max_{t \in G_k } \sum_{n=1}^{t} \sum_{m=1}^{M} \left|\Bar{h}_t^{n}\Bar{w}_t^{mn} - \Bar{h}_{t_{k-1}}^n \Bar{w}_{t_{k-1}}^{mn} \frac{H_{n:t_{k-1}}^{mn}}{H_{n:t}^{mn}} \right|
    \label{eq:groupin-th2-sidek2}
\end{align}
For $G_1$ we have
\begin{equation}
\left|\sum_{t \in G_1} \mathbb{E}[err_t] - \alpha \right| =  \sum_{t \in G_1}  \sum_{n=1}^{t} \sum_{m=1}^{M} \Bar{h}_t^{n} \Bar{w}_t^{mn} (err_t^{mn} - \alpha) \leq \sum_{t \in G_1}  \sum_{n=1}^{t} \sum_{m=1}^{M} \Bar{h}_t^{n} \Bar{w}_t^{m_n} \leq |G_1| \leq \lceil T^{\gamma} \rceil.
\label{eq:groupin-th2-sidek1}
\end{equation}
By summing over all group we have
\begin{align}
    &\left|\sum_{t=1}^T \mathbb{E}[err_t] - \alpha\right| =  \sum_{k=1}^{\lceil T^{1-\gamma} \rceil} \left|\sum_{t \in G_k} \mathbb{E}[err_t] - \alpha \right| \nonumber \\
    &\leq 2T^{\gamma} + \sum_{k=2}^{\lceil T^{1-\gamma} \rceil} \left(\frac{1+ 2\eta}{\eta} \sqrt{T} + 2T^{\gamma} \cdot \max_{t \in G_k } \sum_{n=1}^{t} \sum_{m=1}^{M} \left|\Bar{h}_t^{n}\Bar{w}_t^{mn} - \Bar{h}_{t_{k-1}}^n \Bar{w}_{t_{k-1}}^{mn} \frac{H_{n:t_{k-1}}^{mn}}{H_{n:t}^{mn}} \right| \right) \nonumber \\
    & \leq T^{\frac{3}{2} - \gamma} \frac{1+ 2\eta}{\eta} + 2T^{\gamma} \left(1 +  \sum_{k=2}^{\lceil T^{1-\gamma} \rceil}   \max_{t \in G_k } \sum_{n=1}^{t} \sum_{m=1}^{M} \left|\Bar{h}_t^{n}\Bar{w}_t^{mn} - \Bar{h}_{t_{k-1}}^n \Bar{w}_{t_{k-1}}^{mn} \frac{H_{n:t_{k-1}}^{mn}}{H_{n:t}^{mn}} \right| \right) \nonumber \\
    & \leq \mathcal{O}(T^{\frac{3}{2} - \gamma} + T^{\gamma} \left(1 +  \sum_{k=2}^{\lceil T^{1-\gamma} \rceil}   \max_{t \in G_k } \sum_{n=1}^{t} \sum_{m=1}^{M} \left|\Bar{h}_t^{n}\Bar{w}_t^{mn} - \Bar{h}_{t_{k-1}}^n \Bar{w}_{t_{k-1}}^{mn} \frac{H_{n:t_{k-1}}^{mn}}{H_{n:t}^{mn}} \right| \right)
\end{align}
We define $\beta_{\gamma} (T)$ as:
\begin{equation}
    \beta_{\gamma} (T) := \left(1 +  \sum_{k=2}^{\lceil T^{1-\gamma} \rceil}   \max_{t \in G_k } \sum_{n=1}^{t} \sum_{m=1}^{M} \left|\Bar{h}_t^{n}\Bar{w}_t^{mn} - \Bar{h}_{t_{k-1}}^n \Bar{w}_{t_{k-1}}^{mn} \frac{H_{n:t_{k-1}}^{mn}}{H_{n:t}^{mn}} \right| \right).
    \label{eq:symbol-for-coveragetheorem-alg2}
\end{equation}
Then we have:
\[
\left|\sum_{t=1}^T \mathbb{E}[err_t] - \alpha\right| = \mathcal{O}(T^{\frac{3}{2} - \gamma} + T^{\gamma} \beta_{\gamma} (T)).
\]
\subsection{Proof of Theorem \ref{thm:regret-alg2}, static regret for SAMOCP}
\label{pr:sregret-samocp-proof}
We can write the regret as
\begin{align}
    &\sum_{t \in I_{\Tilde{n}}} \sum_{n \in {\cal A}(t)} \sum_{m=1}^M \Bar{h}_t^{n} \Bar{w}_t^{mn} L(\Bar{\alpha}_t^{mn}, \alpha_{t}^{mn}) - \sum_{t \in I_{\Tilde{n}}}  \sum_{m=1}^M \Bar{w}_t^{m{\Tilde{n}}} L(\Bar{\alpha}_t^{m{\Tilde{n}}}, \alpha_{t}^{m{\Tilde{n}}}) \nonumber \\
    &  + \sum_{t \in I_{\Tilde{n}}}  \sum_{m=1}^M \Bar{w}_t^{m{\Tilde{n}}} L(\Bar{\alpha}_t^{m{\Tilde{n}}}, \alpha_{t}^{m\Tilde{n}}) - \sum_{t \in I_{\Tilde{n}}} L(\Bar{\alpha}_t^{m^*{\Tilde{n}}}, \alpha^{m^*{\Tilde{n}}}),
\end{align}
where $I_{\Tilde{n}}$ denotes the time interval during which expert $\Tilde{n}$ is active, starting at time $t = \Tilde{n}$. The third and fourth terms in the expression are analogous to the regret experienced by expert $\Tilde{n}$, as established in Theorem \ref{thm:regret-alg1}. To evaluate the regret for the first and second terms, we employ Lemma \ref{lem:regret-bound-lem2}. The main difference is in the number of experts considered: For a looser bound, assuming that experts remain active beyond their designated lifetime, the maximum number of experts at each time step $t$ would be $gt$. Thus, we derive the following bound for the first and second terms
\begin{align}
    &\sum_{t \in I_{\Tilde{n}}} \sum_{n \in {\cal A}(t)} \sum_{m=1}^M \Bar{h}_t^{n} \Bar{w}_t^{m{\Tilde{n}}} L(\Bar{\alpha}_t^{mn}, \alpha_{t}^{mn}) - \sum_{t \in I_{\Tilde{n}}}  \sum_{m=1}^M \Bar{w}_t^{m{\Tilde{n}}} L(\Bar{\alpha}_t^{m{\Tilde{n}}}, \alpha_{t}^{m{\Tilde{n}}}) \nonumber \\
    & \leq \frac{\sqrt{|I_{\Tilde{n}}|}\ln{(gt)}}{\sigma} +  \frac{|I_{\Tilde{n}}| \sigma (1+\eta)^2}{\sqrt{|I_{\Tilde{n}}|}} \leq \sqrt{|I_{\Tilde{n}}|}\left(\ln{(gt)} + \sigma^2(1+\eta)^2\right).
    \label{eq:reg-static-saocp-side1}
\end{align}
By summing \eqref{eq:reg-static-saocp-side1} with regret bound of Theorem \ref{thm:regret-alg1} we have
\begin{align}
    &\sum_{t \in I_{\Tilde{n}}} \sum_{n \in {\cal A}(t)} \sum_{m=1}^M \Bar{h}_t^{n} \Bar{w}_t^{mn} L(\Bar{\alpha}_t^{mn}, \alpha_{t}^{mn}) - \sum_{t \in I_{\Tilde{n}}} L(\Bar{\alpha}_t^{m^*{\Tilde{n}}}, \alpha^{m^*{\Tilde{n}}}) \nonumber \\
    & \leq \sqrt{|I_{\Tilde{n}}|}\left(\frac{(1+2\eta)^2}{2\eta} + \frac{\eta}{2\alpha} + \ln{M} + (1+\eta)^2\right) + \sqrt{|I_{\Tilde{n}}|}\left(\ln{(gt)} + \sigma^2(1+\eta)^2\right) \nonumber \\
    & = \sqrt{|I_{\Tilde{n}}|}\left(\frac{(1+2\eta)^2}{2\eta} + \frac{\eta}{2\alpha} + \ln{M} + \ln{g} + (1+\sigma^2)(1+\eta)^2 + \ln{t}\right),
\end{align}
 Until this point, the regret bound we've established applies solely to intervals that start with expert $\Tilde{n}$, where each interval's length corresponds to the lifetime of expert $\Tilde{n}$. However, to derive a regret bound for any arbitrary time interval $I$, we need to partition the interval into subintervals in a suitable manner. As proposed in \citep{daniely2015}, we can divide an interval $I$ into two sequences of non-overlapping and consecutive intervals, denoted as $(I_{-p}, ..., I_0)$ and $(I_1, ..., I_q)$, such that $\frac{|I_{r+1}|}{|I_r|} \leq \frac{1}{2}$ for all $r \in (1, q-1)$, and $\frac{|I_{r}|}{|I_{r+1}|} \leq \frac{1}{2}$ for all $r \in (-p, -1)$. Subsequently, by employing the inequality $\sum_{r=1}^{\infty} \sqrt{2^{-r}T_0} \leq 4\sqrt{T_0}$, and  by replacing $\Tilde{n}$ with $n^*$ using \eqref{eq:optimal-alpha-static-samocp}, we have
\begin{align}
    &\sum_{t \in I} \sum_{n \in {\cal A}(t)} \sum_{m=1}^M \Bar{h}_t^{n} \Bar{w}_t^{mn} L(\Bar{\alpha}_t^{mn}, \alpha_{t}^{mn}) - \sum_{t \in I} L(\Bar{\alpha}_t^{m^*n^*}, \alpha^{m^*n^*}) \nonumber \\
    & = \sum_{r=1}^{q-1}  \sum_{t \in I_{r}} \sum_{n \in {\cal A}(t)} \sum_{m=1}^M \Bar{h}_t^{n} \Bar{w}_t^{mn} L(\Bar{\alpha}_t^{mn}, \alpha_{t}^{mn}) - \sum_{r=1}^{q-1} \sum_{t \in I_{r}} L(\Bar{\alpha}_t^{m^*n^*}, \alpha^{m^*n^*}) \nonumber \\
    & + \sum_{r=-p}^{-1}  \sum_{t \in I_{r}} \sum_{n \in {\cal A}(t)} \sum_{m=1}^M \Bar{h}_t^{n} \Bar{w}_t^{mn} L(\Bar{\alpha}_t^{mn}, \alpha_{t}^{mn}) - \sum_{r=-p}^{-1} \sum_{t \in I_{r}} L(\Bar{\alpha}_t^{m^*n^*}, \alpha^{m^*n^*}) \nonumber \\
    &\leq A\sqrt{|I|} + B\ln{T} \sqrt{|I|}.
    \label{eq:arbit-interval-samocp-stc-reg}
\end{align}
Given that $A$ and $B$ are positive constant variables, we have determined the regret bound for our problem, demonstrating sublinear regret for Algorithm \ref{alg:alg2-expert-selection}.
\subsection{Proof of Lemma \ref{lem:dynamic-regret-samocp}, Dynamic regret for SAMOCP:}
\label{pr:dregret-samocp-proof}
 To prove the regret in a dynamic environment we adopt a method which was first proposed by \citep{besbes2015}. So we can write the dynamic regret in our problem as
 \begin{align}
     & \sum_{t=1}^T \sum_{n \in {\cal A}(t)} \sum_{m=1}^M \Bar{h}_t^{n} \Bar{w}_t^{mn} L(\Bar{\alpha}_t^{mn}, \alpha_{t}^{mn}) - \sum_{t=1}^T L(\Bar{\alpha}_t^{m^*n^*}, \alpha^{m^*n^*}) \nonumber \\
     & + \sum_{t=1}^T L(\Bar{\alpha}_t^{m^*n^*}, \alpha^{m^*n^*})  - \sum_{t=1}^T L(\Bar{\alpha}_t^{m^*n^*}, \alpha_t^{m^*n^*}), 
\label{eq:dynamic-reg-proof}
 \end{align}
where $\alpha_t^{m^*n^*} $ is obtained by \eqref{eq:best-miss-dynamic}. The first two terms in \eqref{eq:dynamic-reg-proof} represent the static regret as defined in Theorem \ref{thm:regret-alg2} and have been shown to be bounded. Consequently, to establish the overall regret bound, it is sufficient to find the upper bounds for the third and fourth terms in \eqref{eq:dynamic-reg-proof}. We begin by dividing the total time interval $T$ into sub-intervals $I_r$ indexed by $r = 1, ..., [T/|I|]$ where $|I|$ is the length of each interval. so we can rewrite \eqref{eq:dynamic-reg-proof} as
\begin{align}
    &  \sum_{r=1}^{[T/|I|]} \sum_{t \in I_r} \sum_{n \in {\cal A}(t)} \sum_{m=1}^M \Bar{h}_t^{n} \Bar{w}_t^{mn} L(\Bar{\alpha}_t^{mn}, \alpha_{t}^{mn}) - \sum_{r=1}^{[T/|I|]} \sum_{t \in I_r} L(\Bar{\alpha}_t^{m^*n^*}, \alpha^{m^*n^*}) \nonumber \\
    &  + \sum_{r=1}^{[T/|I|]} \sum_{t \in I_r} L(\Bar{\alpha}_t^{m^*n^*}, \alpha^{m^*n^*}) - \sum_{r=1}^{[T/|I|]} \sum_{t \in I_r} L(\Bar{\alpha}_t^{m^*n^*}, \alpha_t^{m^*n^*}) .
\end{align}
For the second two terms we have
\[ \sum_{t \in I_r} L(\Bar{\alpha}_t^{m^*n^*}, \alpha^{m^*n^*}) - \sum_{t \in I_r} L(\Bar{\alpha}_t^{m^*n^*}, \alpha_t^{m^*n^*}) 
\leq 2|I|V(L(.)_{t\in I_r}),\]
where $V(L(.)_t)$ is the variability of environment \citep{besbes2015} defined in \eqref{eq:variability-env}. So the regret in \eqref{eq:dynamic-reg-proof} for any arbitrary $|I|$ will be
\begin{align}
    & \sum_{t=1}^T \sum_{n \in {\cal A}(t)} \sum_{m=1}^M \Bar{h}_t^{n} \Bar{w}_t^{mn} L(\Bar{\alpha}_t^{mn}, \alpha_{t}^{mn}) - \sum_{t=1}^T L(\Bar{\alpha}_t^{m^*{n^*}}, \alpha_t^{m^*{n^*}}) \nonumber \\
    &\leq \sum_{r=1}^{[T/|I|]} (A + B\ln{T})\sqrt{|I|} + (2|I|V(L(.)_{t\in I_r})  = (A + B\ln{T})\frac{T}{\sqrt{|I|}} + (2|I|V(L(.)_{t=1}^T)).
\end{align}
Since \eqref{eq:arbit-interval-samocp-stc-reg} holds for any interval $I \subseteq  [T]$, By selecting $|I| = \left| (\frac{T}{V(L(.)_{t=1}^T)})^{\frac{2}{3}} \right|$ we have
\begin{align}
    & \sum_{t=1}^T \sum_{n \in {\cal A}(t)} \sum_{m=1}^M \Bar{h}_t^{n} \Bar{w}_t^{mn} L(\Bar{\alpha}_t^{mn}, \alpha_{t}^{mn}) - \sum_{t=1}^T L(\Bar{\alpha}_t^{m^*{n^*}}, \alpha_t^{m^*{n^*}}) \nonumber \\
    & (A + B\ln{T})T^{\frac{2}{3}} V^{\frac{1}{3}}(L(.)_{t=1}^T) + 2T^{\frac{2}{3}} V^{\frac{1}{3}}(L(.)_{t=1}^T) \leq \Tilde{\mathcal{O}}(T^{\frac{2}{3}} V^{\frac{1}{3}}(L(.)_{t=1}^T) )
\end{align}
\section{Additional Experiments}
\label{sec:add-experimment}
\subsection{Synthetic Dataset}
In this subsection, we present our analysis using synthetic data to compare our proposed method, SAMOCP, with recent adaptive conformal prediction methods designed for dynamic settings. We analyze our experiments with a new set of learning models to demonstrate how our proposed method can effectively utilize a mixture of learning models to cope with distribution shifts in a dynamic environment with unknown distribution changes. Additionally, experiments with synthetic data also confirm that SAMOCP maintains its advantages when varying the number of learning models. Specifically, we conducted experiments for cases with $2$ and $3$ learning models. \\
\textbf{Data Generation:}
To generate synthetic data that mimics real-world scenarios, we employ two distinct transformation sequences. The first transformation sequence introduces visual noise and blur effects through the application of Gaussian blur and random Gaussian noise. This approach aims to subtly degrade image clarity, simulating real-life challenges such as camera focus issues or atmospheric conditions like fog or mist. The Gaussian blur is applied with moderate settings, while random noise is incorporated to simulate sensor noise or digital compression artifacts commonly encountered in digital imagery. The second transformation sequence focuses on color manipulation. By adjusting image attributes such as brightness, contrast, saturation, and hue in minor increments, we challenge the models to perform reliably under varying lighting conditions and color settings—typical variations that occur due to different times of the day or inconsistencies in camera settings. Additionally, a random conversion of some images to grayscale is employed to further challenge the models' dependency on color information. \\
In this experiment, two distinct datasets each containing $3000$ images are generated from each transformation type. These datasets are designed with a fixed number of $20$ classes and hyperparameters \( \xi \) and \( k_{reg} \) are set to $0.1$ and $4$, respectively. The variations between the datasets are due to random elements introduced during image processing, such as differences in which pixels are affected by noise or how color properties are altered. This randomness ensures each dataset contains unique instances, even though they stem from the same transformation principles. By concatenating images from the two datasets, the experiment simulates both gradual and sudden distribution shifts. Gradual shifts are seen within the datasets from a single transformation, while sudden shifts occur when switching between datasets from different transformations.

\begin{table*}[ht]
\centering
\caption{Results on the generated synthetic dataset for Efficientnet\_b0 and GoogLeNet learning models. The target coverage percentage is $90\%$. Bold numbers denote the best results in each column for methods with coverage in the $85-90$ range. Red numbers indicate unacceptable coverage.}
\label{tab:comprehensive-evaluation}
\scriptsize 
\setlength{\tabcolsep}{5pt} 
\begin{tabular}{@{}l >{\raggedright\arraybackslash}p{2cm} cccc c@{}}
\toprule
Model & Method & Coverage (\%) & Avg Width & Avg Regret($\times 10^{-3}$) & Run Time \\
\midrule
 & SAMOCP & 87.90 $\pm$ 0.26 & \textbf{17.54 $\pm$ 0.01} & 0.48 $\pm$ 0.36 & 19.51 $\pm$ 0.03 \\
 & SAOCP(MM) & \textcolor{red}{81.98 $\pm$ 7.35} & 16.54 $\pm$ 1.45 & 6.97 $\pm$ 2.59 & 26.30 $\pm$ 0.24 \\ \\
Efficientnet\_b0 & FACI & 89.80 $\pm$ 0.31 & 17.96 $\pm$ 0.05 & \textbf{0.30 $\pm$ 0.18} & 6.82 $\pm$ 0.02 \\
 & ScaleFreeOGD & \textbf{89.96 $\pm$ 0.00} & 18.05 $\pm$ 0.01 & 1.61 $\pm$ 0.01 & \textbf{2.45 $\pm$ 0.00} \\
 & SAOCP & 88.73 $\pm$ 0.08 & 17.82 $\pm$ 0.01 & 2.18 $\pm$ 0.02 & 34.43 $\pm$ 0.06 \\ \\
GoogLeNet & FACI & 89.66 $\pm$ 0.30 & 18.05 $\pm$ 0.09 & 1.36 $\pm$ 0.79 & 6.89 $\pm$ 0.02 \\
 & ScaleFreeOGD & 89.95 $\pm$ 0.00 & 18.07 $\pm$ 0.03 & 1.82 $\pm$ 0.03 & 2.46 $\pm$ 0.00 \\
 & SAOCP & 88.39 $\pm$ 0.07 & 17.75 $\pm$ 0.02 & 2.55 $\pm$ 0.03 & 34.78 $\pm$ 0.06 \\
\bottomrule
\end{tabular}
\end{table*}

\begin{table*}[ht]
\centering
\caption{Results on the generated synthetic dataset for EfficientNet-B0, DenseNet-121, and GoogLeNet  learning models. The target coverage percentage is $90\%$.Bold numbers denote the best results in each column for methods with coverage in the $85-90$ range. Red numbers indicate unacceptable coverage.}
\label{tab:threemodels-syntheticdata}
\scriptsize 
\setlength{\tabcolsep}{5pt} 
\begin{tabular}{@{}l >{\raggedright\arraybackslash}p{2cm} cccc c@{}}
\toprule
Model & Method & Coverage (\%) & Avg Width & Avg Regret($\times 10^{-3}$) & Run Time \\
\midrule
 & SAMOCP & 88.04 $\pm$ 0.31 & \textbf{17.60 $\pm$ 0.04} & 0.65 $\pm$ 0.45 & 26.39 $\pm$ 0.37 \\
 & SAOCP(MM) & \textcolor{red}{80.95 $\pm$ 7.75} & 16.29 $\pm$ 1.55 & 7.06 $\pm$ 3.03 & 34.80 $\pm$ 0.36 \\ \\
Efficientnet\_b0 & FACI & 89.80 $\pm$ 0.31 & 17.96 $\pm$ 0.05 & \textbf{0.30 $\pm$ 0.18} & 7.37 $\pm$ 0.07 \\
& ScaleFreeOGD & \textbf{89.96 $\pm$ 0.00} & 18.05 $\pm$ 0.01 & 1.61 $\pm$ 0.01 & \textbf{2.65 $\pm$ 0.01} \\
 & SAOCP & 88.73 $\pm$ 0.08 & 17.82 $\pm$ 0.01 & 2.18 $\pm$ 0.02 & 37.25 $\pm$ 0.15 \\ \\
 DenseNet-121 & FACI & 89.72 $\pm$ 0.32 & 18.04 $\pm$ 0.07 & 1.07 $\pm$ 0.73 & 7.42 $\pm$ 0.05 \\
 & ScaleFreeOGD & 89.95 $\pm$ 0.01 & 18.06 $\pm$ 0.02 & 1.81 $\pm$ 0.04 & 2.65 $\pm$ 0.03 \\
 & SAOCP & 88.39 $\pm$ 0.09 & 17.72 $\pm$ 0.01 & 2.53 $\pm$ 0.02 & 37.83 $\pm$ 0.16 \\ \\
GoogLeNet & FACI & 89.66 $\pm$ 0.30 & 18.05 $\pm$ 0.09 & 1.36 $\pm$ 0.79 & 7.49 $\pm$ 0.06 \\
& ScaleFreeOGD & 89.95 $\pm$ 0.00 & 18.07 $\pm$ 0.03 & 1.82 $\pm$ 0.03 & 2.66 $\pm$ 0.03
 \\
 & SAOCP & 88.39 $\pm$ 0.07 & 17.75 $\pm$ 0.02 & 2.55 $\pm$ 0.03 & 37.90 $\pm$ 0.21 \\
\bottomrule
\end{tabular}
\end{table*}
We conducted experiments using a distinct set of learning models. As shown in Table \ref{tab:comprehensive-evaluation}, we incorporated two learning models,  Efficientnet\_b0 and GoogLeNet, where we achieved the smallest prediction set size with coverage close to the target. The SAOCP for  GoogLeNet obtained an average width close to our method; however, it is noteworthy that the maximum number of updates in SAOCP is twice that of SAMOCP, demonstrating that our algorithm achieved this result with lower computational costs. Additionally, its regret is almost five times larger than our method's. We also provide synthetic data analysis using another set of learning models consisting of GoogLeNet, DenseNet-121, and EfficientNet-B0 \citep{tan2019}, as detailed in Table \ref{tab:threemodels-syntheticdata} where our method again was able to construct smaller prediction sets. It should also be noted that, due to severely corrupted data in our synthetic dataset, none of the models were able to produce single width prediction sets that cover true labels.
\subsection{TinyImageNet Dataset}
Here, we conduct the experiment on a new dataset involving a gradual distribution shift using a new real dataset, TinyImageNet-C, which is a corrupted version of the TinyImageNet \citep{le2015} dataset that features $200$ distinct classes. We have also incorporated a new mixture of learning models—GoogLeNet, DenseNet-121, EfficientNet-B0, and MobileNet-V2 \citep{li2021} to demonstrate the performance of our algorithm. In Table \ref{tab:tinyimagenet-grdual}, we detail our proposed method's comparison with previous methods, where, once again, our algorithm is able to achieve smaller prediction set sizes while maintaining coverage close to the target value of $1-\alpha$. It is reasonable to study prediction sets in situations where we achieve coverage close to the desired level. The SAOCP(MM) method achieves coverage of less than $85\%$, which is significantly below the target value. Therefore, we do not consider its prediction sets in our comparison. As demonstrated in the table, SAMOCP obtains smaller prediction sets and operates faster than SAOCP. For the TinyImageNet-C dataset, the parameters  \(\eta\), \( \xi \) and \( k_{reg} \) are set to  \(0.025\), \(0.01\) and \(20\), respectively. Other parameters remain consistent with previous experiments. All real datasets are downloaded from the Zenodo repository.
\begin{table*}[ht]
\centering
\caption{Results on the TinyImageNet-C dataset with a gradual distribution shift. The target coverage is $90\%$, and the average regret is calculated over an interval size of 100. Bold numbers denote the best results in each column for methods with coverage in the 85-90 range. Red numbers indicate unacceptable coverage. SAMOCP achieves the best performance in terms of average width and single width.}
\label{tab:tinyimagenet-grdual}
\scriptsize 
\setlength{\tabcolsep}{5pt} 
\begin{tabular}{@{}llcccccc@{}}
\toprule
Model & Method & Coverage (\%) & Avg Width & Avg Regret($\times 10^{-3}$) & Run Time & Single Width \\
\midrule
& SAMOCP & 87.73 $\pm$ 0.35 & \textbf{171.64 $\pm$ 1.34} & 1.28 $\pm$ 0.20 & 35.22 $\pm$ 0.72 & 0 \\
& SAOCP(MM) & \textcolor{red}{84.91 $\pm$ 1.22} & 165.93 $\pm$ 3.32 & 4.80 $\pm$ 0.77 & 48.55 $\pm$ 0.15 & 0 \\ \\
 GoogLeNet & FACI & 89.69 $\pm$ 0.29 & 177.93 $\pm$ 1.21 & 1.07 $\pm$ 0.67 & 8.48 $\pm$ 0.06 & 0 & \\
 & ScaleFreeOGD & \textbf{89.96 $\pm$ 0.01} & 178.38 $\pm$ 0.49 & 1.41 $\pm$ 0.08 & \textbf{3.16 $\pm$ 0.03} & 0 & \\
 & SAOCP & 88.36 $\pm$ 0.07 & 174.81 $\pm$ 0.50 & 2.02 $\pm$ 0.08 & 40.83 $\pm$ 0.32 & 0 & \\ \\
 
 DenseNet-121 & FACI & 89.68 $\pm$ 0.31 & 176.66 $\pm$ 1.10 & 1.30 $\pm$ 0.74 & 8.54 $\pm$ 0.11 & 0 & \\
 & ScaleFreeOGD & 89.95 $\pm$ 0.02 & 177.24 $\pm$ 0.64 & 1.53 $\pm$ 0.06 & 3.18 $\pm$ 0.05 & 0 & \\
 & SAOCP & 88.48 $\pm$ 0.12 & 173.79 $\pm$ 0.57 & 2.09 $\pm$ 0.04 & 40.09 $\pm$ 0.36 & 0 & \\ \\
Efficientnet\_b0 & FACI & 89.64 $\pm$ 0.35 & 176.78 $\pm$ 1.03 & 1.01 $\pm$ 0.61 & 8.56 $\pm$ 0.07 & 0 & \\
& ScaleFreeOGD & 89.95 $\pm$ 0.01 & 177.23 $\pm$ 0.63 & 1.37 $\pm$ 0.06 & 3.17 $\pm$ 0.05 & 0 & \\
 & SAOCP & 88.30 $\pm$ 0.18 & 173.52 $\pm$ 0.63 & 1.98 $\pm$ 0.08 & 41.06 $\pm$ 0.15 & 0 & \\ \\
Mobilenet\_v2 & FACI & 89.69 $\pm$ 0.30 & 175.26 $\pm$ 0.95 & \textbf{0.98 $\pm$ 0.52} & 8.55 $\pm$ 0.07 & 0 & \\
 & ScaleFreeOGD & 89.95 $\pm$ 0.01 & 175.82 $\pm$ 0.51 & 1.36 $\pm$ 0.07 & 3.19 $\pm$ 0.04 & 0 &\\
 & SAOCP & 88.30 $\pm$ 0.15 & 171.83 $\pm$ 0.55 & 1.94 $\pm$ 0.06 & 40.60 $\pm$ 0.23 & 0 & \\ \\

\bottomrule
\end{tabular}
\end{table*}

\subsection{SAMOCP vs. SAOCP (equal lifetimes)}

We compared our method with a specific version of SAOCP where the hyperparameter $g$ in equation \eqref{eq:lifetime} is set to 8, giving both SAOCP and SAMOCP experts the same lifetime.  We conduct experiments identical to those described in section \ref{sec:experiments}, addressing both sudden and gradual shifts for CIFAR-10C and CIFAR-100C, respectively. Our results are demonstrated in Table \ref{tab:saocp8-cifar10} for CIFAR-10C and in Table \ref{tab:saocp8-cifar100} for CIFAR-100C. For both cases, we observe the advantage of utilizing multiple learning models instead of a single model, as our proposed method, SAMOCP, significantly outperforms SAOCP across all four learning models. SAMOCP achieved better results in terms of coverage, average width, average regret, and single width compared to SAOCP.
 
\begin{table*}[ht]
\centering
\caption{Performance evaluation of SAOCP with same lifetime time as SAMOCP for sudden distribution shift setting on CIFAR-10C.}
\label{tab:saocp8-cifar10}
\scriptsize 
\setlength{\tabcolsep}{5pt} 
\begin{tabular}{@{}l >{\raggedright\arraybackslash}p{2cm} cccc c@{}}
\toprule
Model & Method & Coverage (\%) & Avg Width & Avg Regret($\times 10^{-3}$) & Run Time & Single Width \\
\midrule
 & SAMOCP & \textbf{88.37 $\pm$ 0.23} & \textbf{1.24 $\pm$ 0.06} & \textbf{0.98 $\pm$ 0.11} & 33.75 $\pm$ 0.34 & \textbf{0.69 $\pm$ 0.03} \\ \\
 DenseNet-121 & SAOCP & 87.16 $\pm$ 0.10 & 1.45 $\pm$ 0.03 & 4.36 $\pm$ 0.14 & \textbf{14.73 $\pm$ 0.09} & 0.53 $\pm$ 0.01 \\
ResNet-50 & SAOCP & 87.87 $\pm$ 0.15 & 1.52 $\pm$ 0.01 & 3.91 $\pm$ 0.11 & 14.78 $\pm$ 0.07 & 0.52 $\pm$ 0.01 \\
ResNet-18 & SAOCP & 87.16 $\pm$ 0.18 & 1.51 $\pm$ 0.02 & 4.38 $\pm$ 0.16 & 14.85 $\pm$ 0.08 & 0.51 $\pm$ 0.01 \\
GoogLeNet & SAOCP & 87.74 $\pm$ 0.10 & 1.45 $\pm$ 0.01 & 3.92 $\pm$ 0.15 & 14.84 $\pm$ 0.06 & 0.54 $\pm$ 0.01 \\
\bottomrule
\end{tabular}
\end{table*}

\begin{table*}[ht]
\centering
\caption{Performance evaluation of SAOCP with same lifetime time as SAMOCP for gradual distribution shift setting on CIFAR-100C.}
\label{tab:saocp8-cifar100}
\scriptsize 
\setlength{\tabcolsep}{5pt} 
\begin{tabular}{@{}l >{\raggedright\arraybackslash}p{2cm} cccc c@{}}
\toprule
Model & Method & Coverage (\%) & Avg Width & Avg Regret($\times 10^{-3}$) & Run Time & Single Width \\
\midrule
 & SAMOCP & \textbf{88.16 $\pm$ 0.18} & \textbf{5.43 $\pm$ 0.28} & \textbf{0.92 $\pm$ 0.07} & 34.87 $\pm$ 0.67 & \textbf{0.29 $\pm$ 0.01} \\ \\
 DenseNet-121 & SAOCP & 87.33 $\pm$ 0.14 & 6.12 $\pm$ 0.16 & 4.03 $\pm$ 0.06 & \textbf{14.89 $\pm$ 0.06} & 0.27 $\pm$ 0.00 \\
ResNet-50 & SAOCP & 87.22 $\pm$ 0.12 & 6.75 $\pm$ 0.16 & 3.99 $\pm$ 0.14 & 15.01 $\pm$ 0.07 & 0.26 $\pm$ 0.00\\
ResNet-18 & SAOCP & 87.15 $\pm$ 0.14 & 7.24 $\pm$ 0.22 & 4.10 $\pm$ 0.11 & 15.02 $\pm$ 0.08 & 0.25 $\pm$ 0.00 \\
GoogLeNet & SAOCP & 87.09 $\pm$ 0.16 & 6.78 $\pm$ 0.12 & 4.14 $\pm$ 0.13 & 14.98 $\pm$ 0.09 & 0.26 $\pm$ 0.00 \\
\bottomrule
\end{tabular}
\end{table*}

\subsection{SAMOCP vs. MOCP} 

To validate the advantage of SAMOCP versus MOCP in dynamic environments, we conducted experiments for both sudden and gradual distribution shifts. see Tables \ref{tab:mocp-samocp-cifar10c} and \ref{tab:mocp-samocp-cifar100c}. In both tables, SAMOCP outperforms MOCP in terms of Average Width and Single Width metrics.

\begin{table*}[ht]
\centering
\caption{Comparison of MOCP and SAMOCP on the CIFAR-10C dataset with a sudden distribution shift. The target coverage is 90\%. Bold numbers denote the best results in each column.}
\label{tab:mocp-samocp-cifar10c}
\scriptsize 
\setlength{\tabcolsep}{5pt} 
\begin{tabular}{@{}l >{\raggedright\arraybackslash}p{2cm} ccc@{}}
\toprule
Method & Coverage (\%) & Avg Width & Single Width \\
\midrule
MOCP & \textbf{89.96 $\pm$ 0.33} & 1.29 $\pm$ 0.08 & 0.67 $\pm$ 0.04 \\
SAMOCP & 88.37 $\pm$ 0.23 & \textbf{1.24 $\pm$ 0.06} & \textbf{0.69 $\pm$ 0.03} \\
\bottomrule
\end{tabular}
\end{table*}

\begin{table*}[ht]
\centering
\caption{Comparison of MOCP and SAMOCP on the CIFAR-100C dataset with a gradual distribution shift. The target coverage is 90\%. Bold numbers denote the best results in each column.}
\label{tab:mocp-samocp-cifar100c}
\scriptsize 
\setlength{\tabcolsep}{5pt} 
\begin{tabular}{@{}l >{\raggedright\arraybackslash}p{2cm} ccc@{}}
\toprule
Method & Coverage (\%) & Avg Width & Single Width \\
\midrule
MOCP & \textbf{89.77 $\pm$ 0.19} & 5.85 $\pm$ 0.28 & 0.28 $\pm$ 0.01 \\
SAMOCP & 88.16 $\pm$ 0.18 & \textbf{5.43 $\pm$ 0.28} & \textbf{0.29 $\pm$ 0.01} \\
\bottomrule
\end{tabular}
\end{table*}



\newpage
\section*{NeurIPS Paper Checklist}

\begin{enumerate}

\item {\bf Claims}
    \item[] Question: Do the main claims made in the abstract and introduction accurately reflect the paper's contributions and scope?
    \item[] Answer: \answerYes{} 
    \item[] Justification: We clearly point out our contribution in the abstract and introduction.

\item {\bf Limitations}
    \item[] Question: Does the paper discuss the limitations of the work performed by the authors?
    \item[] Answer: \answerYes{} 
    \item[] Justification: The result tables (sections \ref{sec:experiments} and \ref{sec:add-experimment}) demonstrate that our method does not achieve the best coverage and runs slower than single models.

\item {\bf Theory Assumptions and Proofs}
    \item[] Question: For each theoretical result, does the paper provide the full set of assumptions and a complete (and correct) proof?
    \item[] Answer: \answerYes{} 
    \item[] Justification: For each theorem and lemma in section \ref{sec:model-selection} and \ref{sec:expert-selection} we explained our assumptions and provide clear proof in section \ref{sec:proof}.

    \item {\bf Experimental Result Reproducibility}
    \item[] Question: Does the paper fully disclose all the information needed to reproduce the main experimental results of the paper to the extent that it affects the main claims and/or conclusions of the paper (regardless of whether the code and data are provided or not)?
    \item[] Answer: \answerYes{} 
    \item[] Justification: We provide explanation of our experiment procedure in section \ref{sec:experiments} and also provide full explanation of generating synthetic data set in section \ref{sec:add-experimment}.

\item {\bf Open access to data and code}
    \item[] Question: Does the paper provide open access to the data and code, with sufficient instructions to faithfully reproduce the main experimental results, as described in supplemental material?
    \item[] Answer: \answerYes{} 
    \item[] Justification: A link to the GitHub repository, containing the full implementation of the method described in the paper, is provided in Section \ref{sec:experiments}

\item {\bf Experimental Setting/Details}
    \item[] Question: Does the paper specify all the training and test details (e.g., data splits, hyperparameters, how they were chosen, type of optimizer, etc.) necessary to understand the results?
    \item[] Answer: \answerYes{} 
    \item[] Justification: In sections \ref{sec:experiments} and \ref{sec:add-experimment}, we determine the values of the hyperparameters.

\item {\bf Experiment Statistical Significance}
    \item[] Question: Does the paper report error bars suitably and correctly defined or other appropriate information about the statistical significance of the experiments?
    \item[] Answer: \answerYes{} 
    \item[] Justification: We did every experiment for 10 different random seeds.

\item {\bf Experiments Compute Resources}
    \item[] Question: For each experiment, does the paper provide sufficient information on the computer resources (type of compute workers, memory, time of execution) needed to reproduce the experiments?
    \item[] Answer: \answerYes{} 
    \item[] Justification: In section \ref{sec:add-experimment}, we mentioned that we used an NVIDIA RTX A4000 GPU, and demonstrated the run time for each iteration of our algorithm compared to previous algorithms.
    
\item {\bf Code Of Ethics}
    \item[] Question: Does the research conducted in the paper conform, in every respect, with the NeurIPS Code of Ethics \url{https://neurips.cc/public/EthicsGuidelines}?
    \item[] Answer: \answerYes{} 

\item {\bf Broader Impacts}
    \item[] Question: Does the paper discuss both potential positive societal impacts and negative societal impacts of the work performed?
    \item[] Answer: \answerNA{} 
    \item[] Justification: We do not expect any direct societal impact in our work.
    
\item {\bf Safeguards}
    \item[] Question: Does the paper describe safeguards that have been put in place for responsible release of data or models that have a high risk for misuse (e.g., pretrained language models, image generators, or scraped datasets)?
    \item[] Answer: \answerNA{} 

\item {\bf Licenses for existing assets}
    \item[] Question: Are the creators or original owners of assets (e.g., code, data, models), used in the paper, properly credited and are the license and terms of use explicitly mentioned and properly respected?
    \item[] Answer: \answerYes{} 
    \item[] Justification: In section \ref{sec:experiments} we cited every datasets we employed.

\item {\bf New Assets}
    \item[] Question: Are new assets introduced in the paper well documented and is the documentation provided alongside the assets?
    \item[] Answer: \answerYes{} 
    \item[] Justification: In section \ref{sec:add-experimment} we mentioned that real datasets are downloaded from Zenodo respiratory. 

\item {\bf Crowdsourcing and Research with Human Subjects}
    \item[] Question: For crowdsourcing experiments and research with human subjects, does the paper include the full text of instructions given to participants and screenshots, if applicable, as well as details about compensation (if any)? 
    \item[] Answer: \answerNA{}{} 

\item {\bf Institutional Review Board (IRB) Approvals or Equivalent for Research with Human Subjects}
    \item[] Question: Does the paper describe potential risks incurred by study participants, whether such risks were disclosed to the subjects, and whether Institutional Review Board (IRB) approvals (or an equivalent approval/review based on the requirements of your country or institution) were obtained?
    \item[] Answer: \answerNA{}{} 

\end{enumerate}

\end{document}